\documentclass[journal]{IEEEtran}

\usepackage[pdftex]{graphicx}
\usepackage{amsmath}
\usepackage{amssymb}
\usepackage{booktabs}
\usepackage{tabularx}

\usepackage[noend]{algpseudocode}
\usepackage{algorithm2e}

\usepackage{hyperref}
\hypersetup{
    colorlinks=true,
    linkcolor=black,
    citecolor=black,
    urlcolor=blue,
}
\usepackage{multirow}

\usepackage{tikz}
\usepackage{pgfplots}
\usepackage{caption}
\usepackage{subcaption}
\usepgfplotslibrary{colorbrewer}
\pgfplotsset{cycle list/Set1-9}
\tikzset{every picture/.style={line width=1pt}}
\usetikzlibrary{patterns}
\usepgfplotslibrary{fillbetween}
\usepackage{cite}
\usepackage[eulergreek]{sansmath}
\pgfplotsset{
  tick label style = {font=\sansmath\sffamily\scriptsize},
  every axis label = {font=\sansmath\sffamily\scriptsize},
  y label style={at={(0.05,0.5)}},
  legend style = {font=\sansmath\sffamily\scriptsize},
  label style = {font=\sansmath\sffamily\scriptsize},
}
\DeclareMathAlphabet\mathbfcal{OMS}{cmsy}{b}{n}
\makeatletter
\def\BState{\State\hskip-\ALG@thistlm}
\makeatother

\begin{document}

\title{Deep Unsupervised Contrastive Hashing for Large-Scale Cross-Modal Text-Image Retrieval in Remote Sensing}

\author{Georgii~Mikriukov,~\IEEEmembership{Student Member,~IEEE,}
        Mahdyar~Ravanbakhsh,~\IEEEmembership{Member,~IEEE,}
        and~Begüm~Demir,~\IEEEmembership{Senior~Member,~IEEE}

\thanks{Georgii Mikriukov,
        Mahdyar Ravanbakhsh, and 
        Begüm Demir are with
        Technische Universität Berlin, Berlin, Germany.}}

\markboth{Journal of \LaTeX\ Class Files,~Vol.~13, No.~9, September~2014}%
{Shell \MakeLowercase{\textit{et al.}}: Bare Demo of IEEEtran.cls for Journals}

\maketitle

\begin{abstract}
%HALF PAGE OF MOTIVATION IS TOO MUCH IN ABSTRACT. PLEASE START DIRECTLY WITH CROSS RETRIEVAL AND EXISTING WORKS SUPERVISED AND NOT SCALABLE... THEY ARE NOT BASED ON HASHING I THINK. BUT PLEASE CHECK. PLEASE GIVE MORE EMPHASISE ONWHAT WE PROSE NOVEL FOR SCALABLE AND ACCURATE RETRIEVAL. PLEASE DO NOT PROVIDE MOTIVATION AS: WE PROPOSE THAT TOPIC BECAUUSE IT IS POPULAR, ETC. 
Due to the availability of large-scale multi-modal data (e.g., satellite images acquired by different sensors, text sentences, etc) archives, the development of cross-modal retrieval systems that can search and retrieve semantically relevant data across different modalities based on a query in any modality has attracted great attention in RS. In this paper, we focus our attention on cross-modal text-image retrieval, where queries from one modality (e.g., text) can be matched to archive entries from another (e.g., image). Most of the existing cross-modal text-image retrieval systems require a high number of labeled training samples and also do not allow fast and memory-efficient retrieval due to their intrinsic characteristics. These issues limit the applicability of the existing cross-modal retrieval systems for large-scale applications in RS. To address this problem, in this paper we introduce a novel deep unsupervised cross-modal contrastive hashing (DUCH) method for RS text-image retrieval. The proposed DUCH is made up of two main modules: 1) feature extraction module (which extracts deep representations of the text-image modalities); and 2) hashing module (which learns to generate cross-modal binary hash codes from the extracted representations). Within the hashing module, we introduce a novel multi-objective loss function including: i) contrastive objectives that enable similarity preservation in both intra- and inter-modal similarities; ii) an adversarial objective that is enforced across two modalities for cross-modal representation consistency; iii) binarization objectives for generating representative hash codes. Experimental results show that the proposed DUCH outperforms state-of-the-art unsupervised cross-modal hashing methods on two multi-modal (image and text) benchmark archives in RS. Our code is publicly available at \url{https://git.tu-berlin.de/rsim/duch}.

%\footnote{also available at https://github.com/comrados/duch}.

%Scalable content-based information retrieval is an important task in modern remote sensing (RS) due to ever-increasing amount of satellite data. Hashing is widely used for scalable and memory-efficient retrieval. It allows for the representation of information in common binary space where the similarity between samples can be represented as the hamming distance between hash codes. Cross-modal retrieval aims to learn semantic relationships across different modalities and makes the retrieval process more flexible and versatile. Compared to conventional methods, deep neural networks have a greater capacity to learn semantic relationships between modalities, thus improving the retrieval performance. In this paper we investigate unsupervised deep cross-modal hashing and propose the Deep Unsupervised Cross-Modal Contrastive Hashing (DUCH) method. Firstly, we introduce a novel multi-term contrastive-learning method for cross-modal hashing. The method is supported by adversarial learning, allowing for improved preservation of cross-modal semantic similarity. Secondly, we analyze the influence of data augmentation on each modality and propose a novel rule-based text augmentation method based on semantic embeddings. Experimental results demonstrate that DUCH outperforms state-of-the-art unsupervised cross-modal hashing methods on two RS datasets.

\end{abstract}


\IEEEpeerreviewmaketitle

 \section{Introduction}
 \label{sec:intro}
 
Recent years have witnessed the fast growing of remote sensing (RS) image archives as a result of the advances in satellite technology. As an example, through the Copernicus programme (which is the European flagship satellite initiative in Earth observation), Sentinel satellites reach the scale of more than 10 TB data per day. Accordingly, several methods for fast and accurate search and retrieval in massive archives of RS images have been proposed. Most of the existing methods focus on content based RS image retrieval (CBIR). CBIR systems take a query image and compute the similarity function between the query image and all archive images to find the most similar images to the query \cite{sumbul2021deep}. Recent advances in deep neural networks (DNNs) have led to significant improvement in image retrieval performance due to their high capability to represent complex content of RS images. Most of the existing RS CBIR systems based on DNNs are devoted to: 1) obtain discriminative image descriptors; and 2) achieve scalable search and retrieval. To learn discriminative image descriptors, deep representation learning methods based on a triplet loss function are found very effective for CBIR problems due to their intrinsic characteristic to model similarities of images. These methods exploit image triplets (each of which consists of anchor, positive and negative images), aiming to learn a metric space where the distance between the positive and the anchor images is minimized while that between the negative and anchor images is maximized \cite{GTDRLCBIR, cao2020enhancing, Sumbul2021triplet}. To achieve scalable image retrieval, deep hashing techniques have become a cutting-edge research topic for large-scale RS image retrieval \cite{roy2021hashing, Cheng2021hashing}. These methods map high-dimensional image descriptors into a low-dimensional Hamming space where the image descriptors are described by binary hash codes. Compared with the real-valued features, hash codes allow fast image retrieval through the calculation of the Hamming distances with simple bit-wise XOR operations. In addition, the binary codes can significantly reduce the amount of memory required for storing the content of images. For the detailed survey on RS CBIR methods, we refer the reader to \cite{sumbul2021deep}. 

The above-mentioned methods are defined for single-modality image retrieval problems (called as uni-modal retrieval). For a given query image, uni-modal retrieval systems search for the images with semantically similar contents from the same modality image archive \cite{cao2020enhancing}. However, multi-modal data archives, including different modalities of satellite images as well as textual data, are currently available. Thus, the development of retrieval systems that return a set of semantically relevant results of different modalities given a query in any modality (e.g., using a text sentence to search for RS images) has recently attracted great attention in RS. Due to semantic gap among different modalities, it is challenging to search relevant multi-modal contents among heterogenous data archives. To address this problem, cross-modal retrieval methods that aim to identify relevant data across different modalities are recently introduced in RS \cite{chen2020deep, cheng2021deep, ning2021semantics}. The existing cross-modal retrieval systems in RS are defined based on supervised retrieval techniques. Such techniques require the availability of labeled training samples (i.e., ground reference samples) to be used in the learning phase of the cross-modal retrieval algorithm. The amount and the quality of the available training samples are crucial for achieving accurate cross-modal retrieval. The collection of a sufficient number of reliable labeled samples is time consuming, complex and costly in operational scenarios, and can significantly affect the final accuracy of cross-modal retrieval. Unlike RS, in the computer vision (CV) community, unsupervised and in particular self-supervised cross-modal representation learning methods (which only rely on the alignments between modalities) are widely studied \cite{jung2021contrastive,su2019deep,liu2020joint, chen2020improved,chen2020simple,he2020momentum}. The existing unsupervised contrastive learning methods mainly rely on cross-modal (inter-modality) contrastive objectives to obtain consistent representations across different modalities, while the intra-modal contrastive objectives are ignored. This may lead to learning an inefficient embedding space, where the same semantic content is mapped into different points in the embedding space \cite{zolfaghari2021crossclr}.
 
In this paper, we focus on the cross-modal retrieval between RS image and text sentence (i.e., caption), which consists of two sub-tasks: i) RS image retrieval with a sentence as query; and ii) sentence retrieval with a RS image as query. In detail, we propose a novel deep unsupervised cross-modal contrastive hashing (DUCH) method that considers both inter- and intra-modality contrastive objectives for representation learning. To this end, the proposed DUCH consists of two modules: 1) the feature extraction module that includes two modality-specific encoders to generate the uni-modal deep representation; 2) the hashing module that aims at learning joint feature representations (in terms of binary hash codes) between image and text sentence (i.e., caption) modalities by simultaneously preserving the consistency and modality-invariance in an end-to-end manner. Due to its modules, DUCH ensures scalable and accurate cross-modal retrieval in an unsupervised manner, without requiring any labeled training sample. Experiments carried out on two benchmark archives demonstrate the success of the proposed DUCH compared to state of the art cross-modal hashing methods. The major contributions of this paper can be summarized as follows:
\begin{itemize}
    \item We present an novel unsupervised contrastive hashing method for scalable and accurate cross-modal text-image retrieval in RS.
    \item We introduce a novel multi-objective loss function that consists of: 1) intra- and inter-modal contrastive objectives that enable similarity preservation within and between modalities; 2) an adversarial objective enforcing cross-modal representation consistency; 3) binarization objectives to generate representative binary hash codes.
    \item We propose a rule-based text augmentation method based on word replacement, where words are replaced by semantically similar ones.
    \item We conduct a comparative analysis of data augmentation methods and their influence on cross-modal retrieval performance.
%    \item Experimental results show that the proposed approach outperforms other state-of-the-art unsupervised cross-modal retrieval methods and shrinks the gap between unsupervised and supervised ones.
\end{itemize}

The rest of paper is organized as follows: Section \ref{sec:related} presents the related works, while Section \ref{sec:method} formulates the problem and presents the proposed DUCH. Section \ref{sec:experiments} describes experimental setup and considered datasets, while Section \ref{sec:results} provides experimental results. Finally, Section \ref{sec:conclusion} draws the conclusion of this work.

%\vspace{-0.1 pt}
 \section{Related Work}
 \label{sec:related}

\subsection{Deep Cross-Modal Retrieval}
\label{sec:related-hashing}

Unlike uni-modal retrieval methods (where query and retrieved data share a single modality), cross-modal retrieval methods are designed for scenarios where the queries and retrieval results are from different modalities. In other words, given a query from one modality (e.g., image query), cross-modal retrieval methods provide retrieval results from other modalities (e.g, text, audio). The development of cross-modal retrieval methods are gaining importance in RS due to the availability of large amounts of multi-modal data archives. In \cite{cheng2021deep}, a cross-modal semantic alignment method is presented based on attention and gate mechanisms for text-image cross-modal retrieval. Similarly, for cross-modal text-image retrieval, Yuan et al. \cite{yuan2021exploring} exploit an attention mechanism not only to extract the most representative modality-specific features but also to learn cross-modal features. The method proposed in Ning et al. \cite{ning2021semantics} aims at achieving the representation learning for supervised image-audio retrieval by adding inter-modal (pairwise) consistency and intra-modal consistency loss terms. Given the benefits of its low storage requirements and high retrieval efficiency, hashing has recently received attention also for cross-modal retrieval problems. As an example, Chen et al. \cite{chen2020deep} introduce the deep image–voice retrieval (DIVR) method that considers audio records and RS images as different modalities. DIVR aims at learning consistent binary hash codes across two modalities by using a pairwise similarity loss function. The pairwise similarity objective enforces the similar image–voice features as close as possible. Unlike RS, in the CV community the cross-modal retrieval systems is more extended and widely studied. As an example, Jiang et al. \cite{jiang2017deep} present a deep cross-modal hashing (DCMH) method for learning hash codes from two modalities (i.e., image and text) in an end-to-end manner. DCMH uses a feature learning part and a hash-code learning part to give feedback to each other for learning consistent representations. The inter-modal similarity between modality representations is forced via a pairwise cross-modal loss function. Yang et al. \cite{yang2017pairwise} introduce a similar approach by exploiting both intra- and inter-modal loss terms. Xie et al. \cite{xie2020multi} present a multi-task consistency-preserving adversarial hashing (CPAH) method that uses an attention mechanism to separate representations into modality-specific and modality-common features. CPAH exploits label information to learn consistent modality-specific representations, while an adversarial learning strategy enforces inter-modality semantic consistency. Guo et al. \cite{guo2019jointly} formulate the cross-modal retrieval task as a joint-features classification task via a deep visual-audio network (DVAN). DVAN learns a classification task by concatenating image and audio features. Then, DVAN ranks the likelihood for all the possible joint features in the archive for cross-modal retrieval. The above-mentioned methods learn semantic representations by exploiting pairwise similarities using the label information. Another approach is to learn a metric space by using triplet loss and its different variations. The triplet loss aims at learning a metric space using triplets consisting of anchor, positive and negative samples, where the positive and negative samples are closer and further from the anchor in the feature space, respectively. Wang et al. \cite{wang2017adversarial} introduce a feature learning approach based on a standard triplet loss. Zhang et al. \cite{zhang2018attention} consider attention mechanisms for each modality to learn more representative features from each modality independently from each other. Bai et al. \cite{bai2020deep} improve cross-modal feature consistency by proposing a triplet-based deep adversarial discrete hashing (DADH) method. DADH consists of a hash discriminator and an intermediate feature discriminator, forcing the DADH to learn modality-invariant representations. During the training, the discriminators considered features of one modality to represent ``real'' samples, while features of another modality represent ``fake'' samples. Gu et al. \cite{gu2019adversary} propose to use two modality-specific discriminators to enforce the intra-modal similarity, where each discriminator considers the samples of one modality as ``real'' and the other modality as ``fake''. All the above-mentioned works are supervised methods and require labeled samples for learning the cross-modal representation. Recently, a few unsupervised and self-supervised learning methods are introduced in CV to reduce the need for labeled samples. Li et al. \cite{li2019coupled} propose an unsupervised method based on coupled cycle generative adversarial network (GAN) \cite{goodfellow2014generative}. One network is an outer-cycle GAN that aims at learning intermediate representations from two modalities. The second network is an inner-cycle GAN, which generates hash codes from the intermediate representations. Su et al. \cite{su2019deep} introduce a deep joint-semantics reconstructing hashing (DJSRH) method to learn binary codes that preserve the neighborhood structure in the original data. To this end, DJSRH learns a mapping from different modalities into a joint-semantics affinity matrix. Liu et al. \cite{liu2020joint} propose a joint-modal distribution-based similarity weighting (JDSH) method based on DJSRH, exploiting an additional objective based on cross-modal semantic similarities among samples.

\subsection{Contrastive Learning}
\label{sec:related-contrastive}

Contrastive learning is a metric learning approach that aims to learn a feature space where similar samples are closer and dissimilar samples are farther in the feature space. In the literature, contrastive learning is utilized for a variety of tasks such as image classification \cite{chen2020simple}, object detection \cite{he2020momentum} and retrieval \cite{NEURIPS2020_d89a66c7}. Furthermore, contrastive learning can be used for unsupervised training and supervised fine-tuning with small amounts of the labeled training set \cite{NEURIPS2020_d89a66c7}. Recently, Chen et al. \cite{chen2020simple} introduce a self-supervised contrastive learning framework known as SimCLR for classification of CV images. In SimCLR, different augmentations of an image are considered to represent "positive" samples, while the "negative" samples are randomly chosen from the other samples within a mini-batch. Instead of augmentation in the image space, \cite{he2020momentum} and \cite{chen2020improved} for object detection and image classification problems in CV propose augmentation in the feature space by utilizing two encoders: i) image encoder; and ii) momentum encoder. Feature representations obtained by the image encoder are considered as queries, while those encoded by the momentum encoder are considered as keys. During training, positive pairs are constructed from queries and keys of the current batch, whereas negative pairs are constructed between queries from the current batch and keys of the previous batch. 

Most of the existing contrastive learning methods are designed for single-modal use. However, few works are introduced to adapt contrastive learning for multi-modal data in CV. As an example, Favory et al. \cite{favory2021learning} exploit contrastive learning for self-supervised audio tagging. The model consists of two encoders, one for text tags of audio records and one for the audio. The audio encoder provides the feature representation of the audio spectrogram. Audio embeddings are passed through an attention mechanism and concatenated to obtain a high-level semantic joint-representation, where the cross-modal consistency of semantic representations is enforced with contrastive loss. Zhang et al. \cite{zhang2021cross} introduce cross-modal contrastive learning with distinct intra- and inter-modal terms for text-to-image generation. Semantic feature representations are generated from captions, original images, and generated images. The contrastive loss function has three terms: i) real-generated images intra-modal term; ii) image-text inter-modal term; and iii) word-to-image-regions term. Yuan et al. \cite{yuan2021multimodal} present a contrastive learning method with distinguishable intra- and inter-modal schemes, where the contrastive loss is computed between and within different modalities.

%Contrastive learning methods proposed in CV such as \cite{chen2020simple,he2020momentum,zhang2021cross} provide information on optimal image augmentation methods, which usually include cropping, flipping, rotation, blurring, and color transformations applied for CV images. In addition to the usual augmentation methods for CV images, in RS it is possible to exploit alternative ways to obtain different augmented views. As an example, temporal positive pairs obtained from image patches of the same geographical location in different seasons are used by Manas et al. \cite{manas2021seasonal} for unsupervised training. Ayush et al. \cite{ayush2020geography} in addition to the use of temporal positive pairs reinforce feature encoding by the geo-location classification of the image, where the geo-location is available from the metadata of the image. Stojnic et al. \cite{stojnic2021self} substitute data augmentation by splitting multispectral images into two augmented views containing different subsets of channels as the positive samples. The negative samples are selected from other images in the mini-batch.

\begin{figure*}[htp!]
  \centering
  \includegraphics[width=0.95\linewidth]{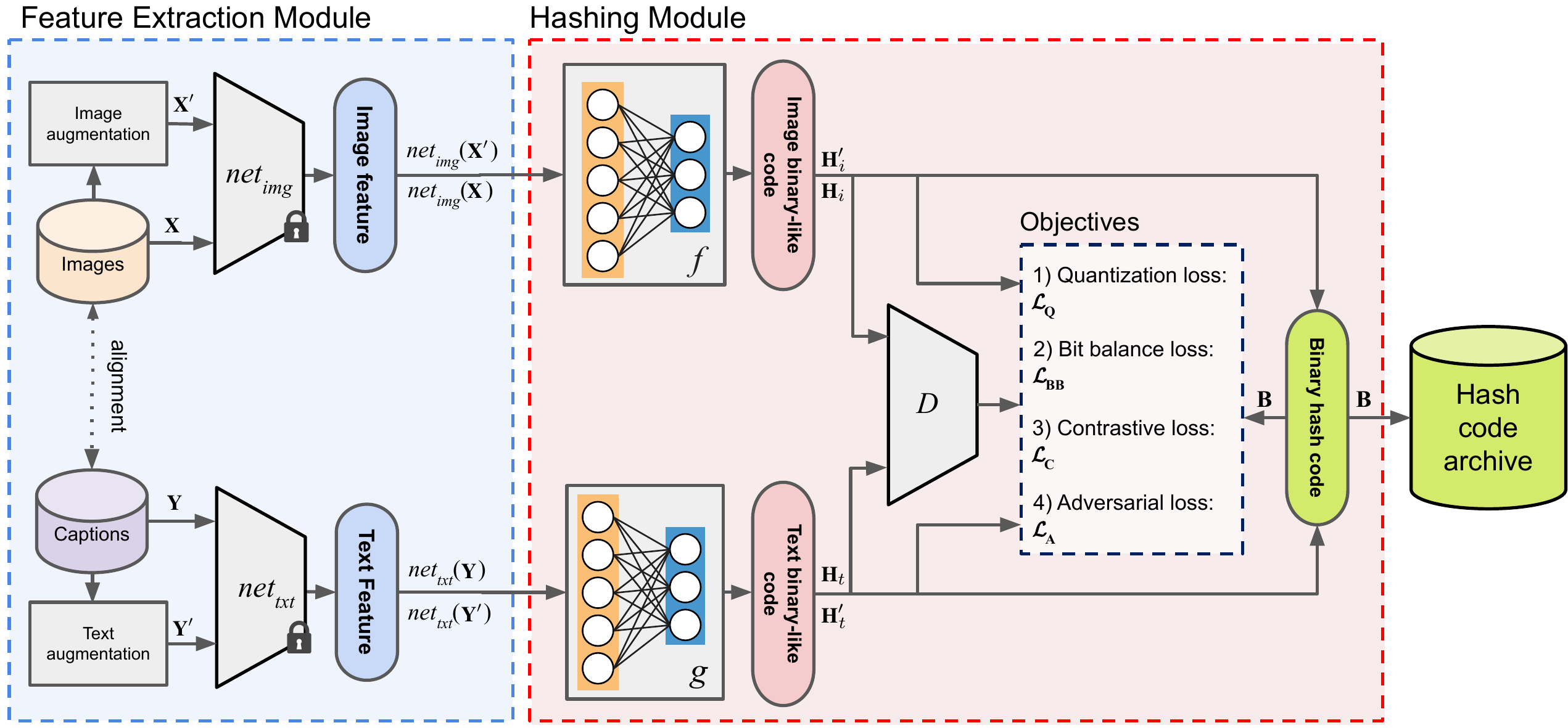}
  \caption{Block diagram of the proposed DUCH unsupervised cross-modal hashing method. Semantic representations are obtained by the modality-specific encoders $net_{img}$ and $net_{txt}$ of the feature extraction module. The hashing module learns two hash functions $f$ and $g$ to generate binary representations from the obtained features. Discriminator $D$ enforces an adversarial objective for the consistency of the representations between two modalities.}
  \label{fig:main-diagram}
\end{figure*}

\subsection{Image and Text Augmentation}
\label{sec:related-augmentation}
 
Data augmentation is crucial to generate different augmented views that can be used in contrastive learning. In case of multi-modal data each modality may need different augmentation methods. Furthermore, in RS some augmentation methods (e.g. color manipulation) are not always a viable choice. Even slight modifications have the potential to distort the information for some tasks. For example, the caption "There is a plane left to an airport terminal" will no longer be semantically relevant to the image if it is flipped vertically. Another example is the excessive color distortion that may lead to errors in the pixel-wise classification of RS images. 

For RS image augmentation in uni-modal contrastive learning, Li et al. \cite{li2021remote} propose two types of augmentations for encouraging the model to learn general spatio-temporal invariance features: 1) spatial transformation by applying random cropping, resizing, flipping, and rotation for the learning of spatial invariance features; and 2) simulating temporal transformations using color distortion, Gaussian blur, random noise to learn temporal invariance features. In \cite{jung2021contrastive} random grayscale, color jitter, and random horizontal flip are used as augmentation methods for metric learning in RS. Heidler et al. \cite{heidler2021self} propose color manipulations, Gaussian blur and rotations to generate augmented views. In addition to the common image augmentation methods (e.g., cropping,  color manipulations, Gaussian blur, etc. ), it is possible to exploit alternative ways to obtain different augmented views in RS. As an example, temporal positive pairs obtained from image patches of the same geographical location in different seasons are used by Manas et al. \cite{manas2021seasonal} for unsupervised training. Ayush et al. \cite{ayush2020geography} in addition to the use of temporal positive pairs reinforce feature encoding by the geo-location classification of the image, where the geo-location is available from the metadata of the image. Stojnic et al. \cite{stojnic2021self} substitute data augmentation by splitting multispectral images into two augmented views containing different subsets of channels as the positive samples. The negative samples are selected from other images in the mini-batch. In order to evaluate the effect of each image augmentation method, Tian et al. \cite{tian2020makes} present an in-depth analysis of data augmentation magnitude influence on a self-supervised contrastive learning. In \cite{tian2020makes}, three types of mutual information sharing are defined between augmented views: i) augmentations that lead to task-related information being discarded by different views; ii) augmentations with the optimal mutual information sharing; and iii) top performance augmentations, which is called as the ``sweet spot''.

To generate augmentation for the text modality three categories of text augmentation methods are distinguished in \cite{feng2021survey}: 1) the rule-based augmentation methods, such as token replacement, swapping, and deletion, which are applied to change input text according to defined rules \cite{wei2019eda}; 2) the interpolation-based augmentations \cite{zhang2017mixup}, which are used to generate an augmented sample from two or more other samples; and 3) the model-based augmentation using language models the aim to generate diverse output sequences from the input sequence (i.e., seq2seq). The back-translation \cite{sennrich2015improving} is one of the seq2seq methods that is used to augment the text by translation from one language to another and then translation back to the original language.

\section{Proposed Method}
\label{sec:method}
\subsection{Problem Definition}
%The cross-modal matching refers to the task of matching (i.e., retrieving) the similar samples from one modality for any given query from a different modality. 
In this paper, we consider an unsupervised cross-modal text-image retrieval problem. Let $\mathcal{D}$ be an unlabelled multi-modal training set that consists of $N$ image and text pairs (i.e., samples), where $\mathcal{D} = \{\textbf{X}, \textbf{Y}\}^N$. In the training set $\mathcal{D}$ each image is described only with one caption (i.e., text modality). $\textbf{X} = \{\textrm{x}_m\}_{m=1}^N$ and $\textbf{Y} = \{\textrm{y}_m\}_{m=1}^N$ are image and text modality sets, respectively, where $\textrm{y}_m$ is the caption for image $\textrm{x}_m$.  %, where $\textrm{x}_m \in \mathbb{R}^{d_i}$ and $\textrm{y}_m \in \mathbb{R}^{d_t}$ are image and text feature vectors. Dimension size of image and text feature vectors are defined by $d_i$ and $d_t$. 
Given the multi-modal training set $\mathcal{D}$, the proposed DUCH aims at learning two hash functions $f(.)$ and $g(.)$ in an unsupervised fashion for image and text modalities, respectively. In detail, the hash functions $f(.)$ and $g(.)$ learn to generate binary hash codes $\textbf{B}_i = f(\textbf{X},\theta_i)$ and $\textbf{B}_t = g(\textbf{Y},\theta_t)$, where $\textbf{B}_i \in \{0 , 1\}^{N \times B}$ and $\textbf{B}_t \in \{0 , 1\}^{N \times B}$ for image and text modalities. $B$ represents the length of binary hash code, while $\theta_i$ and $\theta_t$ are sets of parameters for image and text hashing networks, respectively. In order to learn the hash functions $f(.)$ and $g(.)$, the proposed DUCH includes two main modules:
\begin{enumerate}
    \item feature extraction module that produces deep semantic representations for image and text modalities;
    \item hashing module that generates binary representations from the extracted deep semantic features obtained through the feature extraction module.
\end{enumerate}
Fig. \ref{fig:main-diagram} shows the block diagram of the DUCH, indicating the two modules. In the following sections we describe each module in detail.

\begin{figure*}[t]
  \centering
  \includegraphics[width=\linewidth]{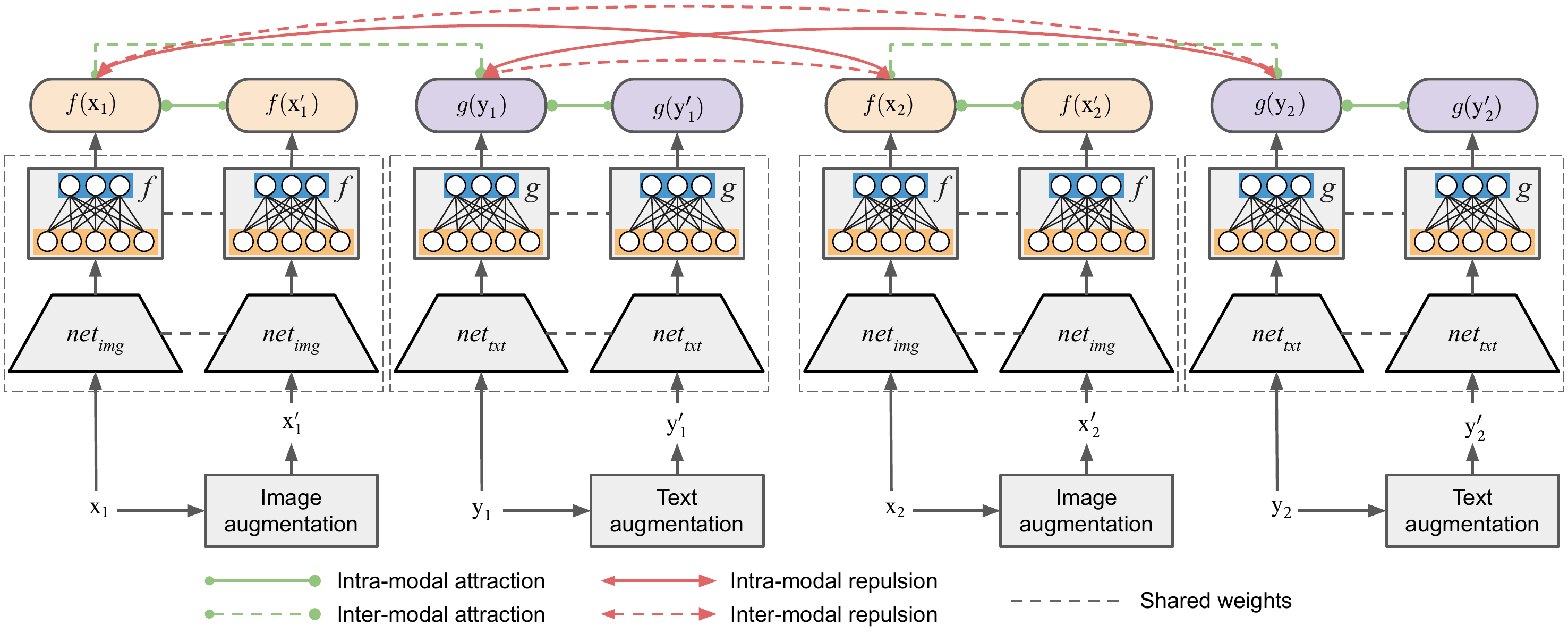}
  \caption{Unsupervised contrastive learning: we consider contrastive objective for both inter-modal (modality-specific) and intra-modal (cross--modal) for image and text modalities.}
  \label{fig:contrastive-learning}
\end{figure*}

\subsection{Feature Extraction Module}
\label{sec:method-components-feature}

The aim of the feature extraction module is to generate deep semantic representations for both images and their captions, and feed the extracted features into the next module (hashing module). To this end, the feature extraction module consists of two networks as modality-specific encoders: 1) an image embedding network denoted as $net_{img}$; 2) a language model for encoding the textual information (i.e., captions) denoted as $net_{txt}$. The blue block in Fig. \ref{fig:main-diagram} shows the feature extraction module including two modality-specific encoders. It is worth noting that the weights of image and text encoders are obtained from pre-trained networks and are frozen during the training of the hashing module. Given the training set $\mathcal{D}$, the image and text embedding are extracted by $net_{img}(\textbf{X})$ and $net_{txt}(\textbf{Y})$, respectively. For the sake of simplicity in the rest of this paper we refer $net_{img}(\textbf{X})$ as $\textbf{X}$, and $net_{txt}(\textbf{Y})$ as $\textbf{Y}$. The unsupervised contrastive representation learning of DUCH requires different views (i.e., augmentations). Therefore, we generate a corresponding augmented set from $\mathcal{D}$, which is defined as $\mathcal{D}^\prime = \{\textbf{X}^\prime, \textbf{Y}^\prime\}^N$, where $\textbf{X}^\prime = \{\textrm{x}_m^\prime\}_{m=1}^N$ and $\textbf{Y}^\prime = \{\textrm{y}_m^\prime\}_{m=1}^N$ are augmented image and caption views with elements $\textrm{x}_m^\prime \in \mathbb{R}^{d_i}$ and $\textrm{y}_m^\prime \in \mathbb{R}^{d_t}$ (for the details of generating augmentations for both modalities see Subsection \ref{sec:experiments-augmentation}). The embeddings of augmented views of images and captions are extracted by $net_{img}$ and $net_{txt}$, respectively. For the sake of simplicity in the rest of this paper we refer $net_{img}(\textbf{X}^\prime)$ as $\textbf{X}^\prime$, and $net_{txt}(\textbf{Y}^\prime)$ as $\textbf{Y}^\prime$. then, the extracted deep semantic representations of the images, captions and the corresponding augmented views are forwarded to the hashing module.

\begin{figure*}[htb]
\centering
  \includegraphics[width=\linewidth]{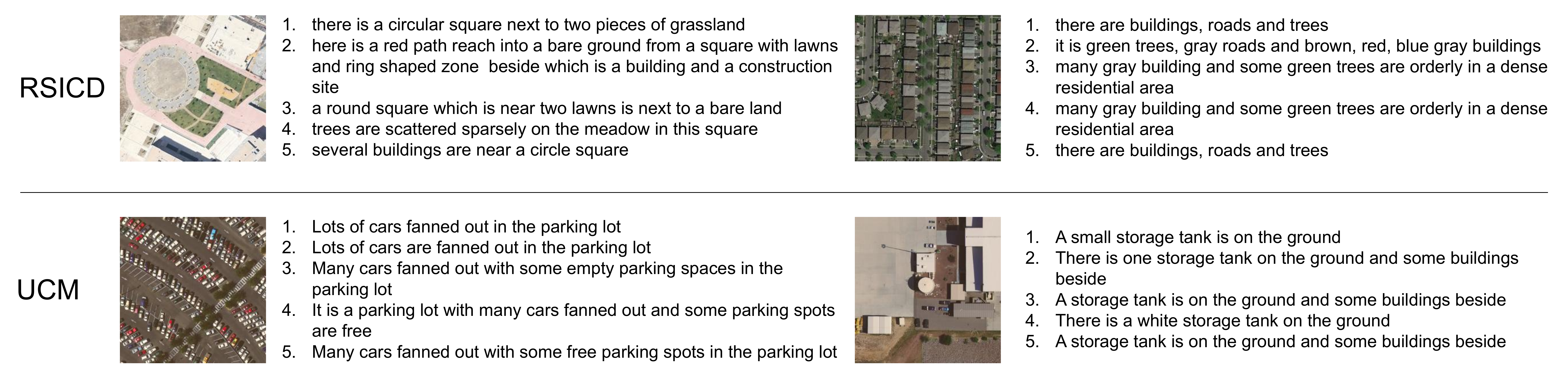}
  \caption{An example of images from the RSICD and UCMerced Land Use archives and the captions associated with them.}
  \label{fig:dataset-samples}
\end{figure*}

\subsection{Hashing Module}
\label{sec:method-components-hash}

The hashing module aims at learning hashing functions $f(.)$ and $g(.)$ to generate binary hash codes from the image and text deep semantic embeddings. To this end, the hashing module of the proposed DUCH includes multiple objectives: 1) contrastive losses; 2) adversarial loss; and 3) binarization losses. The cross-modal contrastive loss is the main objective function for unsupervised representation learning. In detail, we propose to exploit both intra-modal (modality-specific) and inter-modal (cross-modality) contrastive losses to improve the cross-modal representation learning. The modality-invariance capability of the binary codes is ensured by adversarial objective. The binarization objectives are required to approximate better the generated continuous values close to the discrete binary codes. The details of each objective function are discussed in the following.

\noindent{Contrastive objectives:} In the proposed DUCH, we employ the normalized temperature scaled cross-entropy objective function presented in \cite{chen2020simple} for contrastive loss calculation. Furthermore, we consider both inter-modal and intra-modal contrastive losses to ensure better representation learning. The former maps features of both modalities into a common space, while the latter performs the mapping within each modality. The intra-modal and inter-modal contrastive losses are shown in Fig. \ref{fig:contrastive-learning}. The inter-modal contrastive loss $\mathcal{L}_{C_{inter}}$ between image $\textrm{x}_j$ and its paired caption $\textrm{y}_j$ is computed as:

\begin{align}
\label{eq:contrastive-loss-inter}
%\begin{split}
    &\mathcal{L}_{C_{inter}}(\textrm{x}_j,\textrm{y}_j) =\\
    &-\log \frac{S\left( f(\textrm{x}_j), g(\textrm{y}_j) \right) }{ \sum_{k=1, k\neq j}^{M} S\left( f(\textrm{x}_j), f(\textrm{x}_k) \right) + \sum_{k=1}^{M} S\left( f(\textrm{x}_j), g(\textrm{y}_k) \right) }, \nonumber
\end{align}

\noindent where $S(\textrm{u},\textrm{v}) = \exp\left( \cos\left( \textrm{u},\textrm{v} \right) / \tau \right)$, and $\cos\left( \textrm{u}, \textrm{v} \right) = \textrm{u}^T\textrm{v}/\left\| \textrm{u} \right\|\left\| \textrm{v} \right\|$ is the cosine similarity, $\tau$ denotes a temperature, and $M$ is the size of mini-batch. The intra-modal contrastive losses for image and text modalities are computed as: 

\begin{align}
\label{eq:contrastive-loss-intra-img}
    &\mathcal{L}_{C_{img}}(\textrm{x}_j,\textrm{x}^\prime_j) =\\
    &-\log \frac{S\left( f(\textrm{x}_j), f(\textrm{x}^\prime_j) \right) }{ \sum_{k=1, k\neq j}^{M} S\left( f(\textrm{x}_j), f(\textrm{x}_k) \right) + \sum_{k=1}^{M} S\left( f(\textrm{x}_j),f(\textrm{x}^\prime_k) \right) }, \nonumber\\ \nonumber\\
    &\mathcal{L}_{C_{txt}}(\textrm{y}_j,\textrm{y}_j) =\\
    &-\log \frac{S\left( g(\textrm{y}_j), g(\textrm{y}^\prime_j) \right) }{ \sum_{k=1, k\neq j}^{M} S\left( g(\textrm{y}_j), g(\textrm{y}_k) \right) + \sum_{k=1}^{M} S\left( g(\textrm{y}_j),g(\textrm{y}^\prime_k) \right) }, \nonumber
\end{align}

\noindent where $\mathcal{L}_{C_{img}}$ is the contrastive loss between image $\textrm{x}_j$ and its augmentation $\textrm{x}^\prime_j$, while $\mathcal{L}_{C_{txt}}$ is the contrastive loss between caption $\textrm{y}_j$ and its augmentation $\textrm{y}^\prime_j$. The intra-class contrastive losses $\mathcal{L}_{C_{img}}$ and $\mathcal{L}_{C_{txt}}$ along with the cross-modal contrastive loss $\mathcal{L}_{C_{inter}}$ are used to calculate the final contrastive loss. Accordingly, the final contrastive loss $\mathcal{L}_{C}$ is defined as: 

\begin{equation}
    \mathcal{L}_{C} = \mathcal{L}_{C_{inter}} + \lambda_1 \mathcal{L}_{C_{img}} + \lambda_2 \mathcal{L}_{C_{txt}},
    \label{eq:contrastive-loss}
\end{equation}

% adversarial learning for consistency of learned binary representations. discriminator is trained adversarially both on augmented and unaugmented data
\noindent where $\lambda_1$ and $\lambda_2$ are hyperparameters for image and text intra-modal contrastive losses, respectively.

\noindent{Adversarial objective:} To enforce the consistency of representations across image and text modalities, we introduce an adversarial loss within DUCH. In detail, a discriminator network $D$ is trained in an adversarial fashion, where the text binary-like code is assumed to be ``real'' and the image binary-like code is considered as ``fake''. For the discriminator network $D$ an adversarial objective function $\mathcal{L}_{A}$ is calculated as:
\begin{align}
\label{eq:adversarial-loss}
    \mathcal{L}_{A}\left( \textbf{\^X}, \textbf{\^Y} \right) = -\frac{1}{N}\sum^{N}\Big[ &\log\Big( D\big( g \left( \textbf{\^Y} \right), \theta_D \big) \Big) +\\
    &\log \Big( 1 - D\big( f(\textbf{\^X}) , \theta_D\big) \Big) \Big],\nonumber
\end{align}

\noindent where $\textbf{\^X} = \{\textbf{X},\textbf{X}^\prime\}$ denotes the representations for the image modality and $\textbf{\^Y} = \{\textbf{Y},\textbf{Y}^\prime\}$ is the text representations. $D(\textbf{*}, \theta_D)$ $(* =\textbf{\^X}, \textbf{\^Y})$ is the discriminator network $D$ with parameters $\theta_D$. 

\noindent{Binarization objectives:} In order to generate representative hash codes, we consider two binarization objective functions: 1) quantization loss; and 2) bit balance loss. The quantization loss aims to decrease the difference between continuous binary-like codes and discrete hash values. The bit balance loss enforces each output neuron to fire with an equal chance. The use of bit balance loss results in obtaining a binary representation that all the bits of the hash code are used equally. The quantization loss $\mathcal{L}_{Q}$, and bit balance loss $\mathcal{L}_{BB}$ are calculated as: 
\begin{equation}
\label{eq:binarization-quant}
%\begin{split}
    \mathcal{L}_{Q} = \left\| \textbf{B} - \textbf{H}_i \right\|^2_F +
    \left\| \textbf{B} - \textbf{H}_i^\prime \right\|^2_F +
    \left\| \textbf{B} - \textbf{H}_t \right\|^2_F +
    \left\| \textbf{B} - \textbf{H}_t^\prime \right\|^2_F,
%\end{split}
\end{equation}
\begin{equation}
    \mathcal{L}_{BB} = \left\| \textbf{H}_i \cdot 1 \right\|^2_F +
    \left\| \textbf{H}_i^\prime \cdot 1 \right\|^2_F +
    \left\| \textbf{H}_t \cdot 1 \right\|^2_F +
    \left\| \textbf{H}_t^\prime \cdot 1 \right\|^2_F,
    \label{eq:binarization-bit-balance}
\end{equation}
\noindent where $\textbf{H}_i = f( \textbf{X})$, $\textbf{H}_i^\prime = f( \textbf{X}^\prime)$, $\textbf{H}_t = g( \textbf{Y})$, $\textbf{H}_t^\prime = g( \textbf{Y}^\prime)$ are binary like codes for images, augmented images, texts and augmented texts respectively. The final binary code update rule is defined as:
\begin{equation}
        \textbf{B} = sign \left( \frac{1}{2} \left( \frac{\textbf{H}_i + \textbf{H}_i^\prime}{2} + \frac{\textbf{H}_t + \textbf{H}_t^\prime}{2} \right) \right).
    \label{eq:binarization-final-binary-code}
\end{equation}

The overall loss function is a weighted sum of multiple objectives calculated from \eqref{eq:contrastive-loss}, \eqref{eq:adversarial-loss}, \eqref{eq:binarization-quant} and \eqref{eq:binarization-bit-balance}:

\begin{equation}
\small
    \min_{\textbf{B}, \theta_i, \theta_t, \theta_D} \mathcal{L} = \mathcal{L}_C + \alpha \mathcal{L}_{A} + \beta \mathcal{L}_Q + \gamma \mathcal{L}_{BB},
    \label{eq:loss-overall}
\end{equation}

where $\alpha$, $\beta$, $\gamma$ are hyperparameters for adversarial, quantization and bit balance losses terms, respectively. After training the hashing module by optimizing (\ref{eq:loss-overall}), the generated binary hash codes store in the hash code archive for the cross-modal retrieval. In detail, to retrieve semantically similar captions to a query that is provided as an image $\textrm{x}_q$, we compute the Hamming distance between $f(net_{img}(\textrm{x}_q))$ and hash codes in the retrieval archive. The obtained distances are ranked in ascending order, and the top-$k$ captions with the lowest distances are retrieved. Similarly, when the query is provided as a caption $\textrm{y}_q$, the Hamming distance between $g(net_{txt}(\textrm{y}_q))$ and hash codes in the retrieval archive are computed, ranked and top-$K$ images are retrieved.

\section{Dataset description and experimental design}
\label{sec:experiments}

\subsection{Datasets}
\label{sec:experiments-datasets}

We conducted our experiments over two different multi-modal RS archives to evaluate the proposed DUCH. The first dataset is the UC Merced Land Use (denoted as UCMerced) \cite{yang2010bag} archive consists of $2100$ images taken from aerial orthoimagery. We used the captions generated in \cite{qu2016deep} for the images in UCMerced. Images belong to one of $21$ classes (each class contains exactly $100$ images) and has $5$ relevant captions. The second dataset is RSICD \cite{lu2017exploring} archive that includes $10921$ images from $31$ aerial orthoimagery classes. Each image has a size of $224 \times 224$ pixels and has $5$ corresponding captions. Fig. \ref{fig:dataset-samples} shows examples of images together with their associated captions from the UCMerced and RSICD datasets. Since our assumption in the proposed DUCH is that for each image only one caption exists, we used only one randomly selected caption for each image in our experiments. Both archives are split by random selection into three non-overlapping sets: train, query, and retrieval sets (50\%, 10\%, and 40\%, respectively). The training is performed over the training set and evaluation is done using the retrieval set, while the queries are selected from the query set.

\subsection{Data Augmentation}
\label{sec:experiments-augmentation}

For the self-supervised training, we applied different augmentation methods for both modalities. We describe the details of the applied augmentation methods for image and text modalities in the following.

\noindent{Image augmentation:} We applied several image augmentation methods, including Gaussian blur, random rotation, color jittering and cropping. Gaussian blur is applied over images with given kernel size and varying $\sigma$ to obtain different augmentations. In our experiments we applied Gaussian blur with kernel size $3 \times 3$ and $\sigma \in [1.1, 1.3]$. Random rotations were applied in $5$-degree ranges. Because it is required to keep the semantic relations between images and captions, we used small values for degree of rotations (up to $\pm20^\circ$). Flipping or rotating by large values such as $90^\circ$, $180^\circ$ or $270^\circ$ can distort geographic positional relations (e.g., 'above', 'below', 'north', 'south', etc). Color jittering is applied to simulate changes of geographical region as an augmented view for an RS image \cite{li2021remote}. We control color jittering effect via strength parameter. In our experiments, several transforms are applied to a given image simultaneously: $\textit{Brightness}(0.8)$, $\textit{Saturation}(0.8)$, $\textit{Contrast}(0.8)$, $\textit{Hue}(0.2)$. Center and random cropping are common image augmentation methods. In our experiments, center cropping of size $200 \times 200$ is applied to all images after all other augmentations. In case of rotation, augmentation is applied to the image before center cropping. This helps to avoid empty spaces on the borders of the augmented image. As for the random cropping, patches of size $200 \times 200$ are cropped randomly from a give image. We should note that selecting the optimal augmentation depends on the task and modalities. We provide a detailed study for the effect of each augmentation in Section \ref{sec:results-augmentation}. 

\noindent{Text augmentation:} We applied two text augmentation methods to generate augmented captions: back-translation and rule-based augmentation.For the back-translation we consider two approaches. The first approach for back-translation involves translating a caption to a random language from a given set, followed by a translation back to the original language: $\textit{en} \to \textit{random(es, de)} \to \textit{en}$, where $en$, $es$ and $de$ are English, Spanish and German, respectively. The second approach is a chain of back-translation that involves a chain of translations to different languages. The process ends with a translation back to the original language: $\textit{en} \to \textit{es} \to \textit{de} \to \textit{en}$. For back-translation we used pre-trained models proposed in \cite{tiedemann2020opus}.

As rule-based augmentation, we proposed a text augmentation method described in Algorithm \ref{alg:rule-based-augmentation}. Our proposed text augmentation algorithm includes two main steps: 1) replacing original words with relevant acronyms; and 2) measuring the semantic similarity of the augmented sentences with original ones. For the first step, the part-of-speech (POS) tagging method is conducted with Stanford POS tagger \cite{toutanova2003feature}. For the evaluation of token and sentence similarities, GloVe embeddings \cite{pennington2014glove} and BERTScore \cite{zhang2019bertscore} are used, respectively.

\RestyleAlgo{ruled}
\begin{algorithm}
\caption{Rule-based text augmentation.}
\label{alg:rule-based-augmentation}
\textbf{Input:} Captions $\textbf{Y} = \{y_m\}_{m=1}^N$;

\textbf{Output:} Augmented captions $\textbf{Y}^\prime$;

\textbf{Initialization:} Set similarity threshold parameters $\textit{Sim}_{GloVe} = 0.65$ and $\textit{Sim}_{BERTScore} = 0.75$;

\For{each $y_m$}{
    Apply POS-tagging, mark nouns and verbs (\textit{marked tokens}) for replacement;\\
    \For{each marked token}{
        Find nearest words (\textit{replacement candidates}) by distance in \textit{GloVe} embedding;\\
        Exclude \textit{replacement candidates} with \textit{GloVe} similarity lower than $\textit{Sim}_{GloVe}$;\\
        Exclude \textit{replacement candidates}, which are another POS;\\
        Calculate \textit{BERTScore} score between $y_m$ and augmented sentences with \textit{replacement candidates};\\
        Exclude \textit{replacement candidates} with \textit{BERTScore} similarity lower than $\textit{Sim}_{BERTScore}$;\\
        \eIf{replacement candidates $> 0$}{
            Replace \textit{marked token} with \textit{replacement candidate} with highest \textit{BERTScore} $OR$ select a random \textit{replacement candidate};
        }{
            Go to the next \textit{marked token};
        }
        }
}
\end{algorithm}

\subsection{Design of Experiments}
\label{sec:experiments-implementation}

The feature extraction module includes two modality-specific encoders for image and text modalities. In our experiments, for the image modality encoder, a ResNet18 \cite{he2015deep} pretrained on ImageNet is used. We removed the final classification head to extract image feature vectors with the size of $512$. For text feature extraction we employed a pretrained language model model (i.e., \textit{bert-base-uncased} \cite{devlin2019bert}) provided by \textit{Hugging Face} framework. We obtained sentence embeddings of size $768$ by summing four last hidden states for each token in the sentence. During the training phase of the hashing network, the weights of the image and text encoders kept frozen. It is important to mention that the DUCH does not rely on a certain type of image and text encoders architecture, and the modality-specific encoders can be replaced with encoders of different architecture.

Image hashing, text hashing and modality discriminator are fully connected networks with $3$ layers. Table \ref{tab:architecture-modules} shows the architectures of the hashing module blocks. The table uses the following notation: $fc_i$ is the $i$-th fully-connected layer, while $BatchNorm$ represents the batch normalization layer. The hyperparameters $\alpha$, $\beta$, $\gamma$, are selected using a grid search strategy and we set $\alpha = 0.01$, $\beta = 0.001$, $\gamma = 0.01$. The value of $\lambda_1=\lambda_2=1$ by default. The batch size is set to 256. Total number of training epochs is $100$. The initial learning rate is set to $0.0001$, it is decreased decreased by one fifth every $50$ epochs. The Adam optimizer was chosen for $f$ and $g$ networks under weight decay of $0.0005$ and $\beta_1 = 0.9$, $\beta_2 = 0.999$, $\epsilon = 10^{-7}$. For discriminator $D$ the Adam optimizer was chosen, where the weight decay was set to $0.0001$ with $\beta_1 = 0.5$, $\beta_2 = 0.9$, $\epsilon = 10^{-7}$.

\begin{table}[htbp]
    \renewcommand{\arraystretch}{1}
    \setlength{\tabcolsep}{5pt}
    \centering
    \small
    \caption{The architecture specifications of different networks in the hashing module.}
    \label{tab:architecture-modules}
    \begin{tabular}{l|c|c|c}
        \hline
        Module & Layer & Activation & Size \tabularnewline
        \hline\hline
        
        \multirow{4}{*}{Image Hashing Network}
        & $fc_1$ & $ReLU$ & $512$ \\
        \cline{2-4}
        & $fc_2$ & $ReLU$ & $4096$ \\
        \cline{2-4}
        & $BatchNorm$ & - & - \\
        \cline{2-4}
        & $fc_3$ & $tanh$ & $K$ \\
        \hline\hline
        
        \multirow{4}{*}{Text Hashing Network}
        & $fc_1$ & $ReLU$ & $768$ \\
        \cline{2-4}
        & $fc_2$ & $ReLU$ & $4096$ \\
        \cline{2-4}
        & $BatchNorm$ & - & - \\
        \cline{2-4}
        & $fc_3$ & $tanh$ & $K$ \\
        \hline\hline
        
        \multirow{4}{*}{Discriminator}
        & $fc_1$ & $ReLU$ & $K$ \\
        \cline{2-4}
        & $fc_2$ & $ReLU$ & $K \times 2$ \\
        \cline{2-4}
        & $BatchNorm$ & - & - \\
        \cline{2-4}
        & $fc_3$ & - & $K$ \\
        \hline
    \end{tabular}
\end{table}

In order to evaluate the proposed DUCH, we compared it with state-of-the-art supervised and unsupervised methods on UCMerced and RSICD datasets. We select the supervised methods CPAH \cite{xie2020multi}, and the unsupervised methods DJSRH \cite{su2019deep} and JDSH \cite{liu2020joint} (see Section \ref{sec:related} for the details about each method). We trained all models under the same experimental setup for a fair comparison. In our experiments we report the results of both image-to-text retrieval ($I \to T$), and text-to-image retrieval ($T \to I$) in terms of: 1) mean average precision (mAP); and 2) top-$k$ precision curves. Mean Average Precision (mAP) is a commonly used metric for evaluating retrieval performance and defined as follows:

\begin{equation}
    mAP = \frac{1}{N} \sum_{i=1}^{N} \frac{1}{r_i} \sum_{j=1}^{k} P_i(j) \times rel_i(j),
    \label{eq:avaluation-mean-average-precision}
\end{equation}

\noindent where $N$ is the size of the query set, $r_i$ is the number of items that are related to the query sample $i$, and $k$ is the number of samples in the database. In our experiments, the mAP performance was assessed on top-20 retrieved samples (denoted as mAP@20). Top-$k$ precision $P(k)$ is the precision for top $k$-samples sorted by distance from query sample and defined as:

\begin{equation}
    P(k) = \frac{\sum_{i=1}^{k}rel(i)}{k},
    \label{eq:avaluation-top-k-precision}
\end{equation}

\noindent where $rel(i)$ is a sample relevance indicator that equals $1$ if the query and retrieved samples are matched and $0$ otherwise. The Hamming ranking is used for the sorting of hash values. It sorts retrieved samples according to the Hamming distance between query and retrieval samples. The Hamming distance is defined by the number of different bits in two binary codes. Here, larger values correspond to better retrieval. In our experiments to obtain top-$k$ precision curves, $P(k)$ was calculated by varying the number of $k$ retrieved samples in the range of [5-200].

\section{Experimental results}
\label{sec:results}

\begin{figure}[t]
  \centering
  \includegraphics[width=\linewidth]{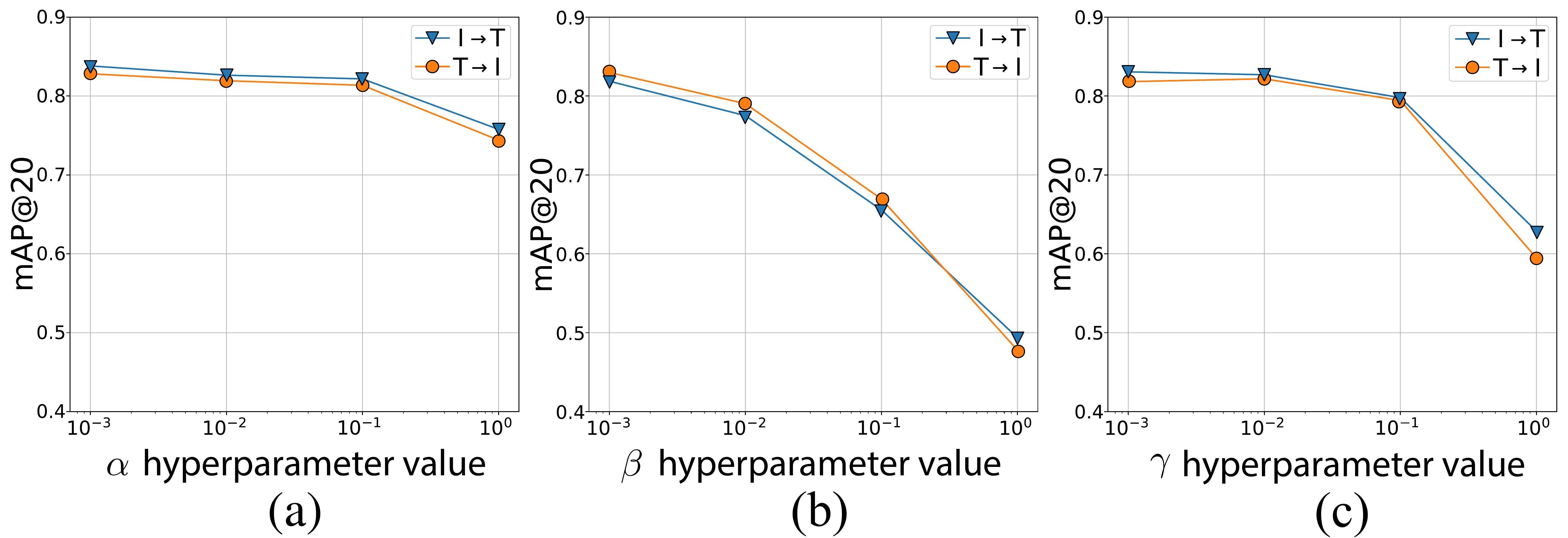}
  \caption{Sensitivity analysis of the proposed DUCH. Image-Text retrieval performance in terms of mAP@20 for $64$ bits hash codes when varying hyperparameters: (a) $\alpha$, (b) $\beta$, and (c) $\gamma$.}
  \label{fig:sensitivity-study}
\end{figure}

\subsection{Sensitivity Analysis of the Proposed Method}
\label{sec:results-sensitivity}

This sub-section presents the sensitivity analysis results for the proposed DUCH in terms of different values for the hyperparameters and the augmentation methods. In detail, we analyzed the DUCH hyperparameters $\alpha$, $\beta$ and $\gamma$, which are the weights for adversarial, quantization and bit balance losses terms, respectively. Fig.\ref{fig:sensitivity-study} shows the results on the RSICD dataset for different values of the hyperparameters $\alpha$, $\beta$ and $\gamma$. The results are obtained on mAP@20 for $B=64$, by varying one hyperparameter, while the values of the other two hyperparameters are set to $0.001$. By assessing the figure, one can observe that different values for $\alpha$ do not significantly affect the performance of the model. This is because the weight for the adversarial loss is always higher or equal to the other loss terms. The adversarial loss plays an important role in enforcing the model to learn consistent representations between the two modalities. If the adversarial loss is dominated by the quantization and bit balance losses terms, the model fails to learn such consistent representations, resulting in a significant performance drop. One can observe that when higher values for the hyperparameter $\beta$ and $\gamma$ are selected, the performance significantly drops. As an example, by increasing $\beta$ from $0.001$ to $0.1$ the mAP@20 for image-text retrieval drops about $20\%$.

\begin{figure}
    \centering
    \begin{subfigure}[b]{0.32\linewidth}
        \centering
        \includegraphics[width=\textwidth]{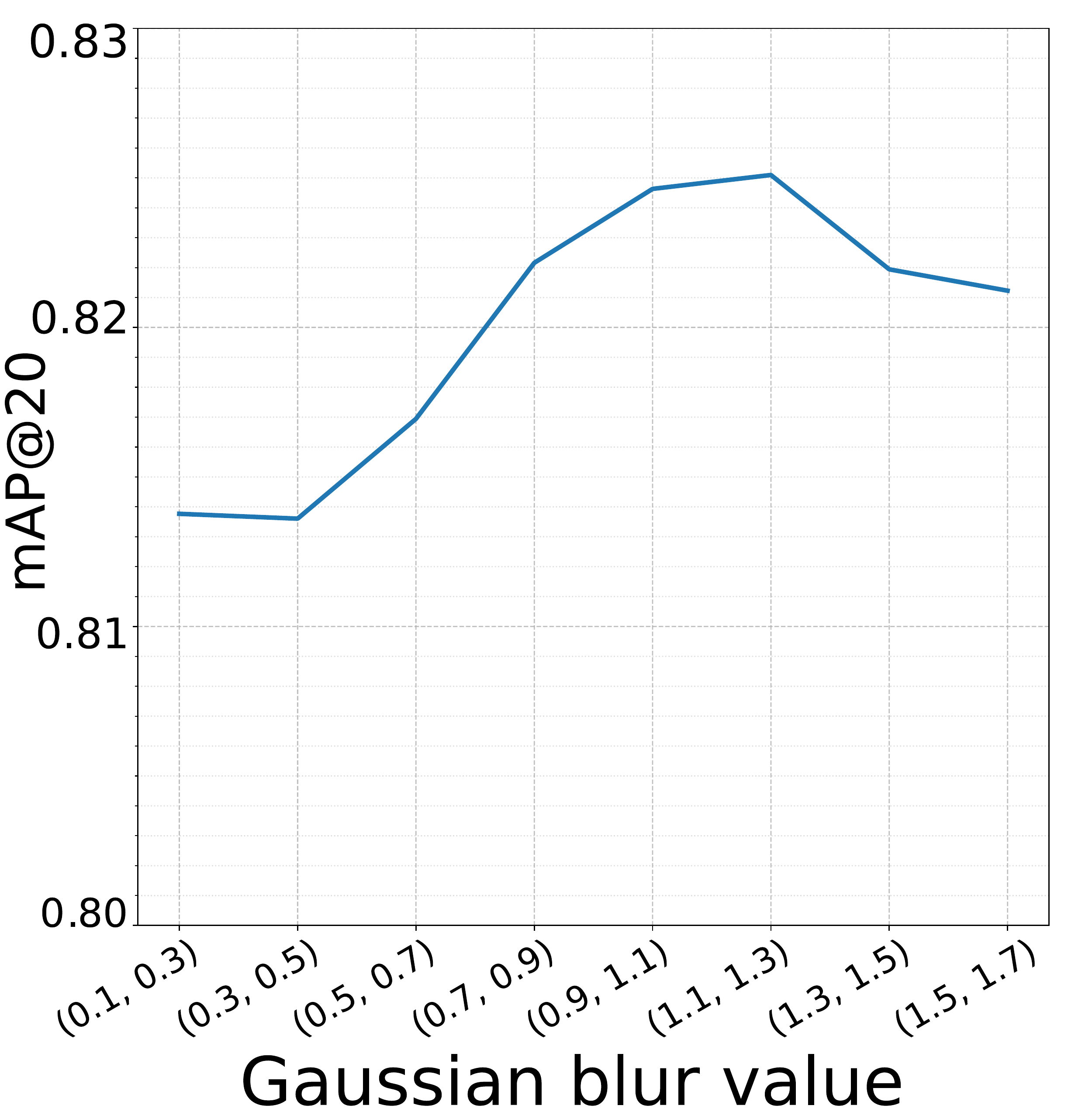}
        \caption[]
        {}    %{\small Gaussian blur with kernel size $3 \times 3$.}
        \label{fig:u-diagram-blur3}
    \end{subfigure}
        \hfill
    \begin{subfigure}[b]{0.32\linewidth}   
        \centering 
        \includegraphics[width=\textwidth]{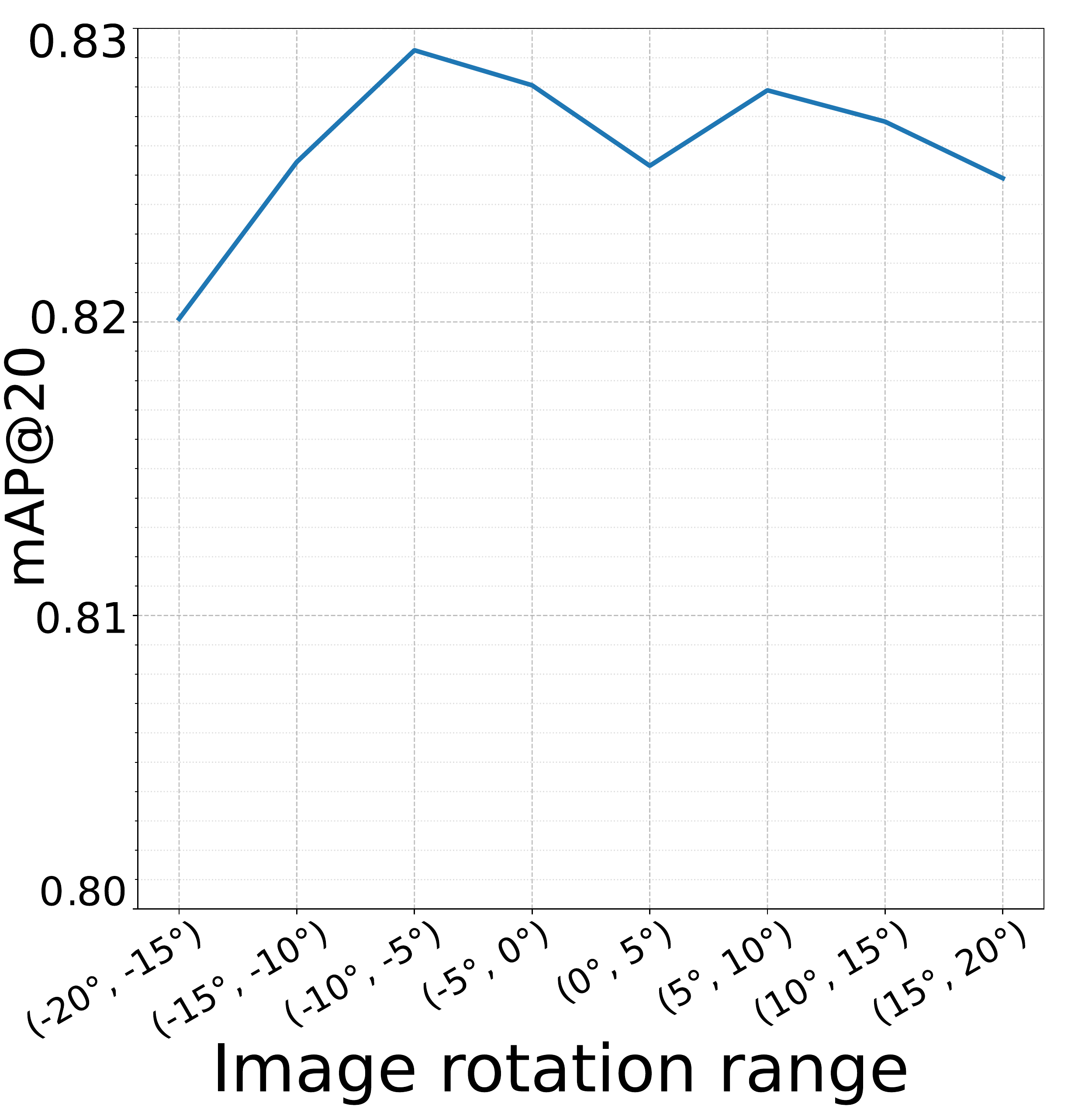}
        \caption[]
        {}    %{\small Random rotation in given range.}
        \label{fig:u-diagram-rotation}
    \end{subfigure}
    \hfill
    \begin{subfigure}[b]{0.32\linewidth}   
        \centering 
        \includegraphics[width=\textwidth]{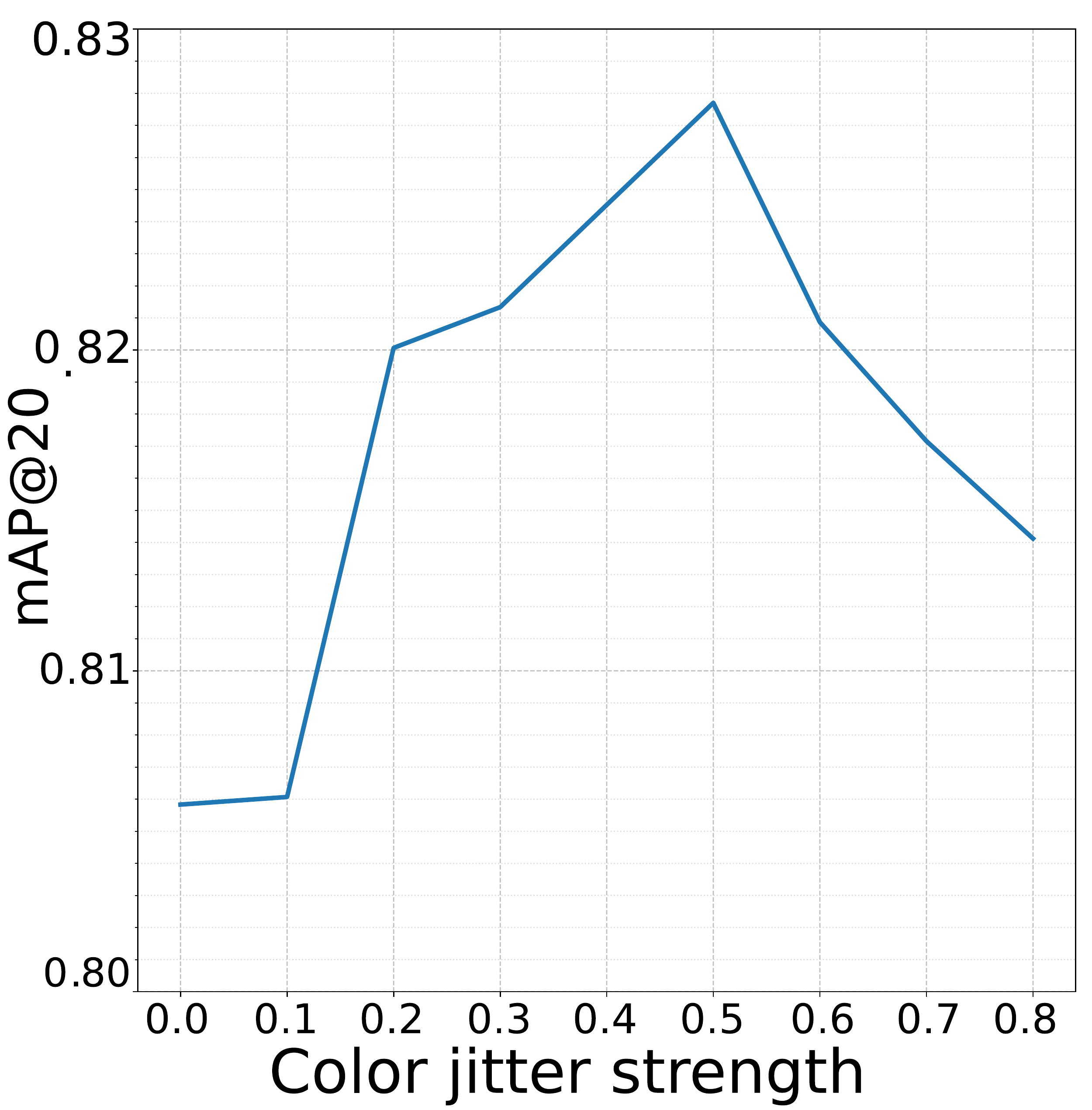}
        \caption[]%
        {}    %{\small Color jittering of given strength.}
        \label{fig:u-diagram-jitter}
    \end{subfigure}
    \caption[]{The reverse U-shape traced out by parameters of augmentation methods: (a) Gaussian blur with kernel size $3 \times 3$, (b) random image rotation, and (c) color jittering.} 
    \label{fig:u-duagrams}
\end{figure}

\subsection{Analysis of the Image and Text Augmentation Methods}
\label{sec:results-augmentation}

In this sub-section, we analyze the effect of different augmentation methods. All the results are presented in terms of mAP@20, when $B=64$. To analyze the effect of the image augmentation methods, we used reverse-U shape diagrams proposed in \cite{tian2020makes}. The reverse-U shape diagrams show how changing the parameter of the individual augmentation method can affect the performance of the proposed DUCH. Fig.\ref{fig:u-duagrams} shows the reverse-U shape diagrams of averaged mAP@20 for image-text retrieval traced out by varying parameters of individual augmentation methods: (a) Gaussian blur; (b) image rotation; and (c) color jittering. The figure shows that selecting small values for the augmentation method hyperparameters (i.e., Gaussian blur, the rotation angle, and color jittering strength) can not increase the performance. On the other hand, selecting higher values for hyperparameters of the augmentation methods could be harmful and affect the retrieval performance. The optimal value for Gaussian blur method is in the range of $[1.1, 1.3]$ with a kernel size of $3 \times 3$. The image rotation yields the best retrieval performance in the range of $[-10^\circ, -5^\circ]$, while the optimal color jittering strength is $0.5$.

Furthermore, we conducted several experiments under different configurations in the augmentation methods to analyze the impact of the text and image augmentations in the context of the proposed contrastive learning. In detail, to analyze the impact of the image augmentation, we considered the following setups:
\begin{itemize}
    \item \textit{NoAugmentation}: Augmentation method is not used, but only center cropping is applied to produce different views for the contrastive objective and duplication of the caption. 
    \item \textit{AugmentedCenterCrop}: A gaussian blur and random rotation are applied to the images, followed by central cropping. %As for the captions, we used \textit{RuleBased} for text augmentation.
    \item \textit{AugmentedRandomCrop}: A gaussian blur and random rotation are applied to the images, followed by random cropping. %The \textit{RuleBased} text augmentation is applied for the captions.
    \item \textit{RandomAugmentation}: One random augmentation (rotation, color jitter, or Gaussian blur) is applied to an image, followed by central cropping. %Text augmentation is done by the \textit{RuleBased} augmentation method.
\end{itemize}
For all the above-mentioned setups, we used \textit{RuleBased} as the text augmentation method. In order to analyze the impact of different text augmentation methods, we considered the following setups:
\begin{itemize}
    \item \textit{RuleBased}: A text is augmented by the rule-based algorithm specified in Algorithm \ref{alg:rule-based-augmentation}. 
    \item \textit{Back-translation}: Generating augmented captions by using the back-translation approach ($\textit{en} \to rabdom(\textit{es} , \textit{de}) \to \textit{en}$) introduced in Section \ref{sec:experiments-augmentation}.
    \item \textit{Back-translationChain}: Generating augmented captions using chain of back-translations ($\textit{en} \to \textit{es} \to \textit{de} \to \textit{en}$) that reviewed in Section \ref{sec:experiments-augmentation}.
\end{itemize}

\begin{table}

% see /images/mAP@20_rsicd_aug_test.png
    \centering
    \small
    \caption{The mAP@20 results on image-to-text ($I \to T$) and text-to-image ($T \to I$) retrieval tasks performed for $B=64$ with different data augmentation setups.}
    \label{tab:augmentation-influence}
    \begin{tabular}{c|c|c|c} 
    \hline
    Modality               & Setup                & $I \to T$ & $T \to I$  \\ 
    \hline\hline
    \multirow{4}{*}{Image} & NoAugmentation       & 0.801 & 0.804 \\
                           & AugmentedCenterCrop  & \textbf{0.836} & \textbf{0.824} \\
                           & AugmentedRandomCrop  & {0.832} & {0.822} \\
                           & RandomAugmentation   & 0.827 & 0.815 \\
    \hline\hline
    \multirow{3}{*}{Text}  & RuleBased            & {0.836} & {0.824} \\
                           & Back-translation      & {0.836} & \textbf{0.829} \\
                           & Back-translationChain & \textbf{0.839} & {0.824} \\
    \hline
    \end{tabular}
\end{table}

For the above-mentioned setups, we only applied \textit{AugmentedCenterCrop} as a method for image augmentation.
The results shown in Table \ref{tab:augmentation-influence} demonstrate the impact of using each augmentation method on the performance of the DUCH. By analyzing the table, one can observe that the \textit{AugmentedCenterCrop} augmentation method is the most effective one for image augmentation, while \textit{AugmentedRandomCrop} performs just marginally worse. This is due to a partial loss of important details since RS images contain complex information and the random crop can exclude important information. For the text augmentation methods the \textit{RuleBased}, \textit{Back-translation} and \textit{Back-translationChain} methods show marginal differences. However, the \textit{Back-translation} is computationally costly, depends on the number of intermediate languages. Furthermore, increasing the number of intermediate languages may lead to significant distortions in the meaning of a sentence. Hence, we used the \textit{RuleBased} method for text augmentations in our experiments.

\begin{table}
\centering
\small
\caption{The mAP@20 results for image-to-text ($I \to T$) and text-to-image ($T \to I$) retrieval tasks under different configurations of the proposed DUCH, where $B=64$.}
\label{tab:ablation-study}
\def\arraystretch{1.05}
\begin{tabular}{p{1.75cm}|p{3.45cm}|c|c} 
\hline
Method & Configuration & $I \to T$ & $T \to I$  \\ 
\hline\hline
DUCH         & standard (See Section \ref{sec:experiments}) & 0.836 & 0.824 \\
\hline\hline
DUCH-NA      & excluding $\mathcal{L}_{adv}$: $\alpha = 0$  & 0.831 & 0.822 \\
DUCH-NQ      & excluding $\mathcal{L}_{Q}$: $\beta = 0$     & 0.818 & 0.800 \\
DUCH-NB      & excluding $\mathcal{L}_{BB}$: $\gamma = 0$   & 0.828 & 0.826 \\
\hline\hline
DUCH-CL      & excluding $\mathcal{L}_{C_{img}}$, $\mathcal{L}_{C_{txt}}$: $\lambda_1 = 0$, $\lambda_2 = 0$    & 0.758 & 0.765 \\
DUCH-CL-I    & excluding $\mathcal{L}_{C_{txt}}$: $\lambda_2 = 0$   & 0.811 & 0.796 \\
DUCH-CL-T    & excluding $\mathcal{L}_{C_{img}}$: $\lambda_1 = 0$   & 0.813 & 0.815 \\
\hline
\end{tabular}
\end{table}

\subsection{Ablation Study}
\label{sec:results-ablation}

% see file /images/ablation_plots_rsicd.png

To analyze the influence of each objective, we designed different configurations by excluding individual objectives from DUCH. In detail, we compared the standard DUCH including (which includes all the proposed objectives) with its different versions defined on the basis of six different configurations:

\begin{itemize}
    \item DUCH-NA: The model without discriminator and adversarial learning, $\alpha = 0$ in Eq. \eqref{eq:loss-overall}.
    \item DUCH-NQ: The model when the quantization loss is removed from overall loss function, $\beta = 0$ in Eq. \eqref{eq:loss-overall}.
    \item DUCH-NB: The model when the bit balance loss is removed from overall loss function, $\gamma = 0$ in Eq. \eqref{eq:loss-overall}.
    \item DUCH-CL: The model when intra-model contrastive loss terms are ignored: $\lambda_1 = 0$, $\lambda_2 = 0$ in Eq. \eqref{eq:contrastive-loss}.
    \item DUCH-CL-I: The model when intra-modal contrastive loss term for the text modality is ignored: $\lambda_1 = 1$, $\lambda_2 = 0$ in Eq. \eqref{eq:contrastive-loss}.
    \item DUCH-CL-T: The model when intra-modal contrastive loss term for the image modality is ignored: $\lambda_1 = 0$, $\lambda_2 = 1$ in Eq. \eqref{eq:contrastive-loss}.
\end{itemize}

The results of ablation study are given in Table \ref{tab:ablation-study}. We reported mAP@20 on image-to-text ($I \to T$) and text-to-image ($T \to I$) retrieval tasks obtained when $B=64$. From the table, one can observe that the controllable objectives $\mathcal{L}_{adv}$, $\mathcal{L}_Q$ and $\mathcal{L}_{BB}$, from \eqref{eq:loss-overall} marginally contribute to the overall performance. The highest performance drop was observed in the case of DUCH-NQ. By removing the quantization objective, the performance drops about $2\%$. Considering DUCH-NA and DUCH-NB configurations, we observe that adversarial and bit balance losses have a less significant impact on the $I \to T$ and $T \to I$ retrieval tasks. This means that the proposed unsupervised DUCH is able to learn consistent representations across two modalities. The most important observation is the impact of the intra-modal contrastive losses. Using intra-modal contrastive losses effectively boosts cross-modal unsupervised training compared to individual intra-modal contrastive loss in case of DUCH-CL. Without the intra-modal losses, the performance significantly drops (about $8\%$). This shows the importance of the joint use of intra-modal losses to improve the inter-modal contrastive loss. Furthermore, by comparing DUCH-CL-I and DUCH-CL-T, one can observe that the text intra-modal contrastive objective is more influential on the overall performance with respect to the image intra-modal objective. This is mainly due to using a modality-specific encoder for the images pre-trained on a different domain (i.e., CV images), resulting in the image embedding that is less representative for RS images.

\begin{table}
% see /data/out_table.csv
\setlength{\tabcolsep}{4pt}
\centering
\small
\caption{The mAP@20 results on image-to-text ($I \to T$) and text-to-image ($T \to I$) retrieval tasks for different values of $B$ for the UCMerced datasets. ``S'' and ``U'' represent the learning type of the considered methods as supervised and unsupervised, respectively.}
\label{tab:retrieval-performance-ucm}
\begin{tabular}{c|lc|cccc} 
\hline
% header
Task & Method && $B$=16 & $B$=32 & $B$=64 & $B$=128  \\ 
\hline\hline
% I2T
\multirow{4}{*}{$I\to T$}        

 & CPAH \cite{xie2020multi}&S        & 0.706 & \textbf{0.802} & \textbf{0.891} & \textbf{0.914} \\
% & DADH \cite{bai2020deep}          & 0.891 & 0.921 & 0.921 & 0.919 \\
%\cline{2-6}

 & DJSRH \cite{su2019deep}&U       & 0.686 & 0.711 & 0.735 & 0.754 \\
 & JDSH \cite{liu2020joint}&U     & 0.462 & 0.751 & 0.820 & 0.829 \\
 & DUCH \textit{(proposed)}&U  & \textbf{0.760} & 0.794 & 0.844 & 0.870 \\
\hline\hline

% T2I
\multirow{4}{*}{$T\to I$}

 & CPAH \cite{xie2020multi}&S        & 0.782 & \textbf{0.891} & \textbf{0.987} & \textbf{0.982} \\
% & DADH \cite{bai2020deep}      & \textbf{0.940} & \textbf{0.981} & \underline{0.977} & \underline{0.964} \\
%\cline{2-6}

 & DJSRH \cite{su2019deep}&U       & 0.738 & 0.755 & 0.776 & 0.800 \\
 & JDSH \cite{liu2020joint}&U     & 0.509 & 0.794 & 0.884 & 0.904 \\
 & DUCH \textit{(proposed)}&U     & \textbf{0.799} & 0.851 & 0.916 & 0.927 \\
\hline
\end{tabular}
\end{table}

\subsection{Comparison of the Proposed Method with Different Cross-Modal Retrieval Methods}
\label{sec:results-retrieval}

In this sub-section, we evaluate the effectiveness of the proposed DUCH method compared to different supervised and unsupervised cross-modal retrieval methods, which are: the supervised method CPAH \cite{xie2020multi}, and the unsupervised methods DJSRH \cite{su2019deep} and JDSH \cite{liu2020joint} (see Section \ref{sec:related} for the details about each method). We evaluated the performance of the two tasks image-to-text and text-to-image retrieval in terms of mAP@20 on the RSICD and the UCMerced datasets when $B={16,32,64,128}$.

\begin{table}
% see /data/out_table.csv
\setlength{\tabcolsep}{4pt}
\centering
\small
\caption{The mAP@20 results on image-to-text ($I \to T$) and text-to-image ($T \to I$) retrieval tasks for different values of $B$ for the RSCID datasets. ``S'' and ``U'' represent the learning type of the considered methods as supervised and unsupervised, respectively.}
\label{tab:retrieval-performance-rsicd}
\begin{tabular}{c|lc|cccc} 
\hline
% header
Task & Method && $B$=16 & $B$=32 & $B$=64 & $B$=128  \\ 
\hline\hline
% I2T
\multirow{4}{*}{$I\to T$}        

 & CPAH \cite{xie2020multi} &S       & 0.428 & 0.587 & 0.636 & 0.696 \\
% & DADH \cite{bai2020deep}          & \textbf{0.696} & 0.703 & 0.723 & 0.712 \\
%\cline{2-6}

 & DJSRH \cite{su2019deep}&U       & 0.411 & 0.665 & 0.688 & 0.722  \\
 & JDSH \cite{liu2020joint}&U     & 0.385 & 0.720 & 0.796 & 0.815 \\
 & DUCH \textit{(proposed)}&U  & \textbf{0.684} & \textbf{0.791} & \textbf{0.836} & \textbf{0.829} \\
\hline\hline

% T2I
\multirow{4}{*}{$T\to I$}

 & CPAH \cite{xie2020multi}&S        & 0.452 & 0.598 & 0.667 & 0.706  \\
% & DADH \cite{bai2020deep}      & \textbf{0.713} & \underline{0.771} & 0.744 & 0.739 \\
%\cline{2-6}

 & DJSRH \cite{su2019deep}&U       & 0.422 & 0.685 & 0.705 & 0.733  \\
 & JDSH \cite{liu2020joint}&U     & 0.418 & 0.751 & 0.799 & {0.815}  \\
 & DUCH \textit{(proposed)}&U     & \textbf{0.697} & \textbf{0.780} & \textbf{0.824} & \textbf{0.826} \\
\hline
\end{tabular}
\end{table}

\begin{figure}
  \centering
  \includegraphics[width=\linewidth]{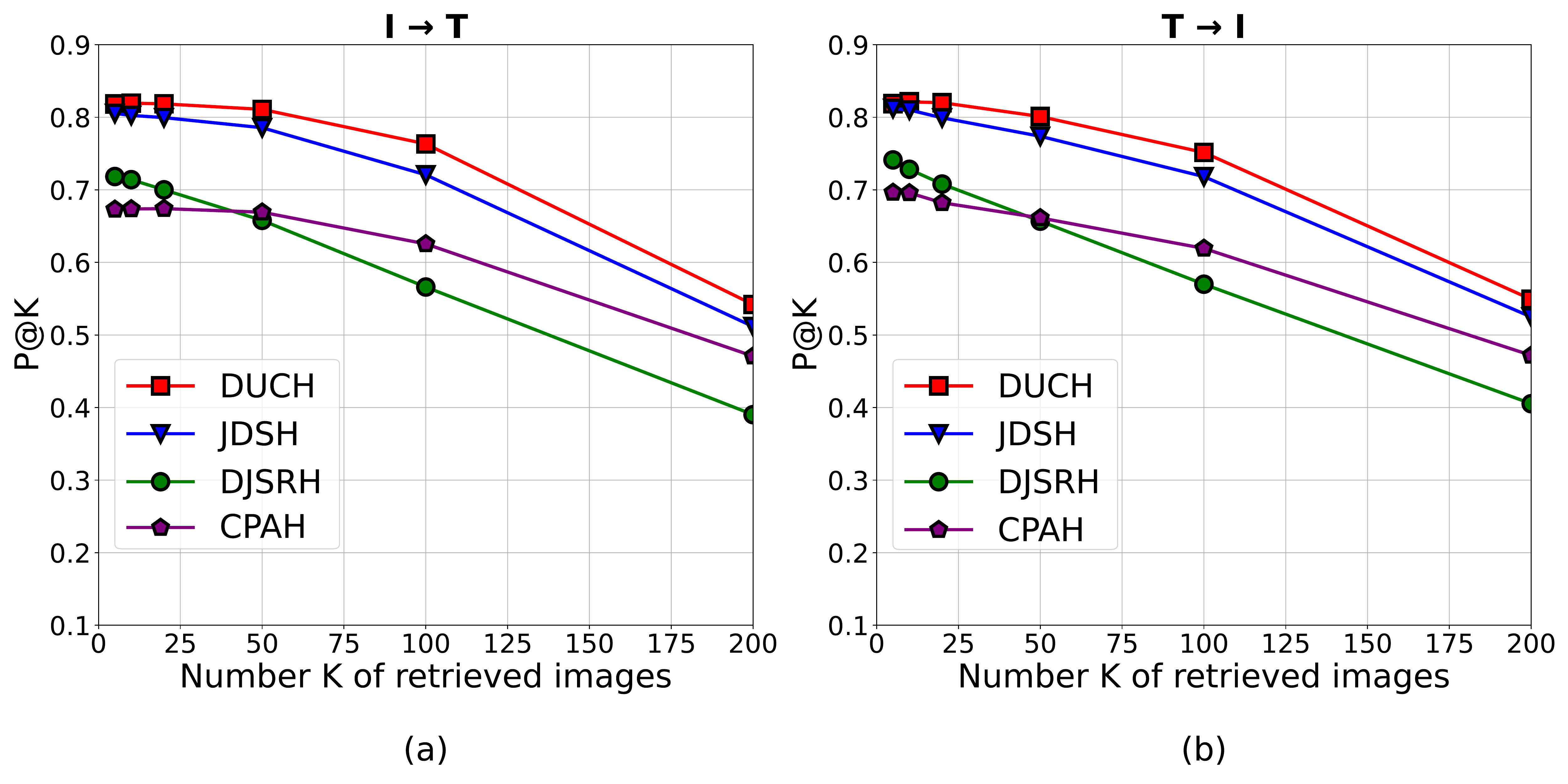}
  \caption{Precision versus number of retrieved images (P@$K$) obtained when $B$=64 for: (a) image-to-text, (b) text-to-image retrieval for the RSICD dataset.}
  \label{fig:precision-at-k}
\end{figure}

\begin{figure*}[t!]
    \newcommand{\figwidth}{0.09\linewidth}
    \newcommand{\figheight}{0.65in}
    
    \definecolor{wrong}{HTML}{cc0000}
    \definecolor{correct}{HTML}{2d862d}
    \definecolor{neutral}{HTML}{595959}
    
    \fboxsep=0pt%padding thickness
    \fboxrule=1pt%border thickness

    \begin{minipage}[c]{\figwidth}
        \centering
        \centerline{Method}\medskip
    \end{minipage}
    \begin{minipage}[c]{\figwidth}
        \centering
        \centerline{Image Query}\medskip
    \end{minipage}
    \begin{minipage}[c]{\figwidth}
        \centering
        \centerline{Caption}\medskip
    \end{minipage}
    \begin{minipage}[c]{\figwidth}
        \centering
        \centerline{1\textsuperscript{st}}\medskip
    \end{minipage}
    \begin{minipage}[c]{\figwidth}
        \centering
        \centerline{2\textsuperscript{nd}}\medskip
    \end{minipage}    
    \begin{minipage}[c]{\figwidth}
        \centering
        \centerline{3\textsuperscript{rd}}\medskip
    \end{minipage}
    \begin{minipage}[c]{\figwidth}
        \centering
        \centerline{4\textsuperscript{th}}\medskip
    \end{minipage}
    \begin{minipage}[c]{\figwidth}
        \centering
        \centerline{5\textsuperscript{th}}\medskip
    \end{minipage}
    \begin{minipage}[c]{\figwidth}
        \centering
        \centerline{10\textsuperscript{th}}\medskip
    \end{minipage}
    \begin{minipage}[c]{\figwidth}
        \centering
        \centerline{20\textsuperscript{th}}\medskip
    \end{minipage}

%CPAH I2T
    \begin{minipage}[c]{\figwidth}
        \centering
        \centerline{
        \parbox[t][0.6in][c]{0.6in}{CPAH \cite{xie2020multi}}
        }
    \end{minipage}
    \begin{minipage}[c]{\figwidth}
        \centering
        \fcolorbox{neutral}{neutral}{\centerline{\includegraphics[height=\figheight]{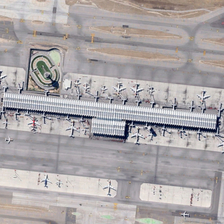}}}\medskip
    \end{minipage}
    \begin{minipage}[c]{\figwidth}
        \tiny
        \color{neutral}
        \centerline{
        \parbox[t][0.6in][t]{0.6in}{{Many planes are in an airport surrounded by many buildings}}
        }
    \end{minipage}
   \begin{minipage}[c]{\figwidth}
        \tiny
        \centerline{
        \color{correct}
        \parbox[t][0.6in][t]{0.6in}{A lot of airplanes are parked in this complex airport surrounded by farmland}
        }
    \end{minipage}
    \begin{minipage}[c]{\figwidth}
        \tiny
        \centerline{
        \color{correct}
        \parbox[t][0.6in][t]{0.6in}{Many planes are in an airport surrounded by many buildings}
        }
    \end{minipage}    
    \begin{minipage}[c]{\figwidth}
        \tiny
        \centerline{
        \color{wrong}
        \parbox[t][0.6in][t]{0.6in}{The white church is located in the center of the square}
        }
    \end{minipage}
    \begin{minipage}[c]{\figwidth}
        \tiny
        \centerline{
        \color{correct}
        \parbox[t][0.6in][t]{0.6in}{Many planes are in an airport near some green trees}
        }
    \end{minipage}
    \begin{minipage}[c]{\figwidth}
        \tiny
        \centerline{
        \color{correct}
        \parbox[t][0.6in][t]{0.6in}{Some planes are parked in an airport surrounded by green meadows and some trees}
        }
    \end{minipage}
    \begin{minipage}[c]{\figwidth}
        \tiny
        \centerline{
        \color{wrong}
        \parbox[t][0.6in][t]{0.6in}{A church with several green trees is near a road with many cars}
        }
    \end{minipage}
    \begin{minipage}[c]{\figwidth}
        \tiny
        \centerline{
        \color{correct}
        \parbox[t][0.6in][t]{0.6in}{Many planes are parked in an airport near some storage tanks}
        }
    \end{minipage}

%DJSRH I2T
    \begin{minipage}[c]{\figwidth}
        \centering
        \centerline{
        \parbox[t][0.6in][c]{0.6in}{DJSRH \cite{su2019deep}}
        }
    \end{minipage}
    \begin{minipage}[c]{\figwidth}
        \centering
        \fcolorbox{neutral}{neutral}{\centerline{\includegraphics[height=\figheight]{images/retrievals/airport_1.jpg}}}\medskip
    \end{minipage}
    \begin{minipage}[c]{\figwidth}
        \tiny
        \color{neutral}
        \centerline{
        \parbox[t][0.6in][t]{0.6in}{Many planes are parked near a building with parking lots in an airport}
        }
    \end{minipage}
   \begin{minipage}[c]{\figwidth}
        \tiny
        \centerline{
        \color{wrong}
        \parbox[t][0.6in][t]{0.6in}{Pieces of bare yellow land }
        }
    \end{minipage}
    \begin{minipage}[c]{\figwidth}
        \tiny
        \centerline{
        \color{correct}
        \parbox[t][0.6in][t]{0.6in}{Many planes are parked in an airport}
        }
    \end{minipage}    
    \begin{minipage}[c]{\figwidth}
        \tiny
        \centerline{
        \color{correct}
        \parbox[t][0.6in][t]{0.6in}{Several planes are parked in an airport surrounded by meadows}
        }
    \end{minipage}
    \begin{minipage}[c]{\figwidth}
        \tiny
        \centerline{
        \color{correct}
        \parbox[t][0.6in][t]{0.6in}{Four planes are parked in an airport near several buildings with parking lots}
        }
    \end{minipage}
    \begin{minipage}[c]{\figwidth}
        \tiny
        \centerline{
        \color{correct}
        \parbox[t][0.6in][t]{0.6in}{Many planes are parked in an airport near several runways and some green trees}
        }
    \end{minipage}
    \begin{minipage}[c]{\figwidth}
        \tiny
        \centerline{
        \color{wrong}
        \parbox[t][0.6in][t]{0.6in}{Yellow beach is near a piece of green ocean with a line of white wave}
        }
    \end{minipage}
    \begin{minipage}[c]{\figwidth}
        \tiny
        \centerline{
        \color{wrong}
        \parbox[t][0.6in][t]{0.6in}{Many buildings are around a rectangular church}
        }
    \end{minipage}

%JDSH I2T
    \begin{minipage}[c]{\figwidth}
        \centering
        \centerline{
        \parbox[t][0.6in][c]{0.6in}{JDSH \cite{liu2020joint}}
        }
    \end{minipage}
    \begin{minipage}[c]{\figwidth}
        \centering
        \fcolorbox{neutral}{neutral}{\centerline{\includegraphics[height=\figheight]{images/retrievals/airport_1.jpg}}}\medskip
    \end{minipage}
    \begin{minipage}[c]{\figwidth}
        \tiny
        \color{neutral}
        \centerline{
        \parbox[t][0.6in][t]{0.6in}{Many planes are parked in an airport near several runways and some green trees}
        }
    \end{minipage}
   \begin{minipage}[c]{\figwidth}
        \tiny
        \centerline{
        \color{correct}
        \parbox[t][0.6in][t]{0.6in}{Some planes are parked in an airport surrounded by green meadows and some trees}
        }
    \end{minipage}
    \begin{minipage}[c]{\figwidth}
        \tiny
        \centerline{
        \color{correct}
        \parbox[t][0.6in][t]{0.6in}{Some planes are parked in an airport}
        }
    \end{minipage}    
    \begin{minipage}[c]{\figwidth}
        \tiny
        \centerline{
        \color{correct}
        \parbox[t][0.6in][t]{0.6in}{Five planes are in an airport}
        }
    \end{minipage}
    \begin{minipage}[c]{\figwidth}
        \tiny
        \centerline{
        \color{correct}
        \parbox[t][0.6in][t]{0.6in}{Many planes are parked in an airport surrounded by green trees}
        }
    \end{minipage}
    \begin{minipage}[c]{\figwidth}
        \tiny
        \centerline{
        \color{correct}
        \parbox[t][0.6in][t]{0.6in}{Several planes are in an airport near some buildings and green trees }
        }
    \end{minipage}
    \begin{minipage}[c]{\figwidth}
        \tiny
        \centerline{
        \color{correct}
        \parbox[t][0.6in][t]{0.6in}{Many planes are parked in an airport near several runways and some green trees}
        }
    \end{minipage}
    \begin{minipage}[c]{\figwidth}
        \tiny
        \centerline{
        \color{correct}
        \parbox[t][0.6in][t]{0.6in}{Many planes are in an airport near some buildings and many green trees}
        }
    \end{minipage}

%DUCH I2T
    \begin{minipage}[c]{\figwidth}
        \centering
        \centerline{
        \parbox[t][0.6in][c]{0.6in}{DUCH \small\textit{(proposed)}}
        }
    \end{minipage}
    \begin{minipage}[c]{\figwidth}
        \centering
        \fcolorbox{neutral}{neutral}{\centerline{\includegraphics[height=\figheight]{images/retrievals/airport_1.jpg}}}\medskip
    \end{minipage}
    \begin{minipage}[c]{\figwidth}
        \tiny
        \color{neutral}
        \centerline{
        \parbox[t][0.6in][t]{0.6in}{Many planes are parked near a building with parking lots in an airport} %
        }
    \end{minipage}
   \begin{minipage}[c]{\figwidth}
        \tiny
        \centerline{
        \color{correct}
        \parbox[t][0.6in][t]{0.6in}{Many planes are in an airport surrounded by many buildings}
        }
    \end{minipage}
    \begin{minipage}[c]{\figwidth}
        \tiny
        \centerline{
        \color{correct}
        \parbox[t][0.6in][t]{0.6in}{Several planes are parked in an airport surrounded by meadows}
        }
    \end{minipage}    
    \begin{minipage}[c]{\figwidth}
        \tiny
        \centerline{
        \color{correct}
        \parbox[t][0.6in][t]{0.6in}{Some planes are parked in an airport }
        }
    \end{minipage}
    \begin{minipage}[c]{\figwidth}
        \tiny
        \centerline{
        \color{correct}
        \parbox[t][0.6in][t]{0.6in}{Many planes are in an airport near some buildings and many green trees }
        }
    \end{minipage}
    \begin{minipage}[c]{\figwidth}
        \tiny
        \centerline{
        \color{correct}
        \parbox[t][0.6in][t]{0.6in}{Some planes are parked in an airport surrounded by green meadows and some trees }
        }
    \end{minipage}
    \begin{minipage}[c]{\figwidth}
        \tiny
        \centerline{
        \color{correct}
        \parbox[t][0.6in][t]{0.6in}{Many planes are in an airport }
        }
    \end{minipage}
    \begin{minipage}[c]{\figwidth}
        \tiny
        \centerline{
        \color{correct}
        \parbox[t][0.6in][t]{0.6in}{Some planes are parked in an airport near runway }
        }
    \end{minipage}

    \caption{$I \to T$ retrieval results obtained when $B=64$ for RSICD dataset.}
    \label{fig:retrieval_i2t}
\end{figure*}

\begin{figure*}[t!]
    \newcommand{\figwidth}{0.09\linewidth}
    \newcommand{\figheight}{0.65in}
    
    \fboxsep=0pt%padding thickness
    \fboxrule=1pt%border thickness
    \definecolor{wrong}{HTML}{cc0000}
    \definecolor{correct}{HTML}{2d862d}
    \definecolor{neutral}{HTML}{595959}

    \begin{minipage}[c]{\figwidth}
        \centering
        \centerline{Method}\medskip
    \end{minipage}
    \begin{minipage}[c]{\figwidth}
        \centering
        \centerline{Text Query}\medskip
    \end{minipage}
    \begin{minipage}[c]{\figwidth}
        \centering
        \centerline{Image}\medskip
    \end{minipage}
    \begin{minipage}[c]{\figwidth}
        \centering
        \centerline{1\textsuperscript{st}}\medskip
    \end{minipage}
    \begin{minipage}[c]{\figwidth}
        \centering
        \centerline{2\textsuperscript{nd}}\medskip
    \end{minipage}    
    \begin{minipage}[c]{\figwidth}
        \centering
        \centerline{3\textsuperscript{rd}}\medskip
    \end{minipage}
    \begin{minipage}[c]{\figwidth}
        \centering
        \centerline{4\textsuperscript{th}}\medskip
    \end{minipage}
    \begin{minipage}[c]{\figwidth}
        \centering
        \centerline{5\textsuperscript{th}}\medskip
    \end{minipage}
    \begin{minipage}[c]{\figwidth}
        \centering
        \centerline{10\textsuperscript{th}}\medskip
    \end{minipage}
    \begin{minipage}[c]{\figwidth}
        \centering
        \centerline{20\textsuperscript{th}}\medskip
    \end{minipage}

%CPAH T2I
    \begin{minipage}[c]{\figwidth}
        \centering
        \centerline{
        \parbox[t][0.6in][c]{0.6in}{CPAH \cite{xie2020multi}}
        }
    \end{minipage}
    \begin{minipage}[c]{\figwidth}
        \tiny
        \color{neutral}
        \centerline{
        \parbox[t][0.6in][t]{0.6in}{The wasteland has both striped and radial texture}
        }
    \end{minipage}
    \begin{minipage}[c]{\figwidth}
        \centering
         \fcolorbox{neutral}{neutral}{\centerline{\includegraphics[height=\figheight]{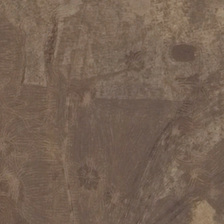}}}\medskip
    \end{minipage}
    \begin{minipage}[c]{\figwidth}
        \centering
        \fcolorbox{wrong}{wrong}{\centerline{\includegraphics[height=\figheight]{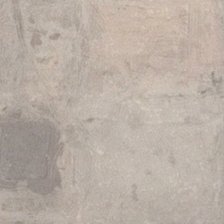}}}\medskip
    \end{minipage}
    \begin{minipage}[c]{\figwidth}
        \centering
        \fcolorbox{correct}{correct}{\centerline{\includegraphics[height=\figheight]{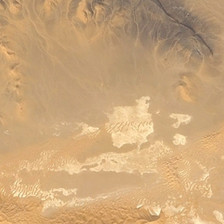}}}\medskip
    \end{minipage} 
    \begin{minipage}[c]{\figwidth}
        \centering
        \fcolorbox{correct}{correct}{\centerline{\includegraphics[height=\figheight]{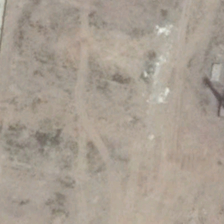}}}\medskip
    \end{minipage}
    \begin{minipage}[c]{\figwidth}
        \centering
        \fcolorbox{correct}{correct}{\centerline{\includegraphics[height=\figheight]{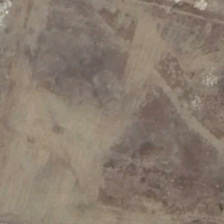}}}\medskip
    \end{minipage}
    \begin{minipage}[c]{\figwidth}
        \centering
        \fcolorbox{wrong}{wrong}{\centerline{\includegraphics[height=\figheight]{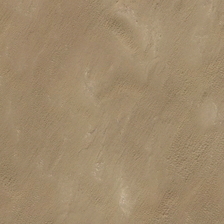}}}\medskip
    \end{minipage}
    \begin{minipage}[c]{\figwidth}
        \centering
        \fcolorbox{wrong}{wrong}{\centerline{\includegraphics[height=\figheight]{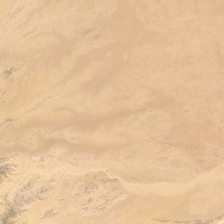}}}\medskip
    \end{minipage}
    \begin{minipage}[c]{\figwidth}
        \centering
        \fcolorbox{wrong}{wrong}{\centerline{\includegraphics[height=\figheight]{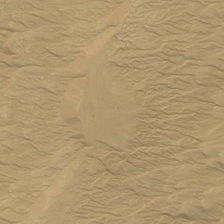}}}\medskip
    \end{minipage}

%DJSRH T2I
    \begin{minipage}[c]{\figwidth}
        \centering
        \centerline{
        \parbox[t][0.6in][c]{0.6in}{DJSRH \cite{su2019deep}}
        }
    \end{minipage}
    \begin{minipage}[c]{\figwidth}
        \tiny
        \color{neutral}
        \centerline{
        \parbox[t][0.6in][t]{0.6in}{The wasteland has both striped and radial texture}
        }
    \end{minipage}
    \begin{minipage}[c]{\figwidth}
        \centering
         \fcolorbox{neutral}{neutral}{\centerline{\includegraphics[height=\figheight]{images/retrievals/bareland_45.jpg}}}\medskip
    \end{minipage}
    \begin{minipage}[c]{\figwidth}
        \centering
        \fcolorbox{correct}{correct}{\centerline{\includegraphics[height=\figheight]{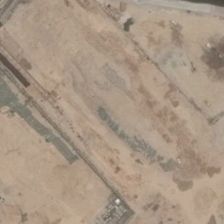}}}\medskip
    \end{minipage}
    \begin{minipage}[c]{\figwidth}
        \centering
        \fcolorbox{correct}{correct}{\centerline{\includegraphics[height=\figheight]{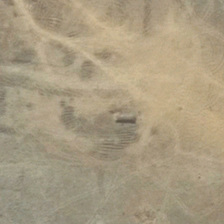}}}\medskip
    \end{minipage} 
    \begin{minipage}[c]{\figwidth}
        \centering
        \fcolorbox{wrong}{wrong}{\centerline{\includegraphics[height=\figheight]{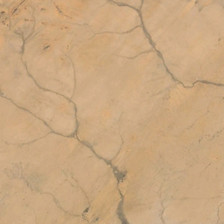}}}\medskip
    \end{minipage}
    \begin{minipage}[c]{\figwidth}
        \centering
        \fcolorbox{correct}{correct}{\centerline{\includegraphics[height=\figheight]{images/retrievals/bareland_174.jpg}}}\medskip
    \end{minipage}
    \begin{minipage}[c]{\figwidth}
        \centering
        \fcolorbox{correct}{correct}{\centerline{\includegraphics[height=\figheight]{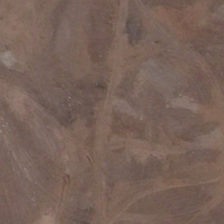}}}\medskip
    \end{minipage}
    \begin{minipage}[c]{\figwidth}
        \centering
        \fcolorbox{correct}{correct}{\centerline{\includegraphics[height=\figheight]{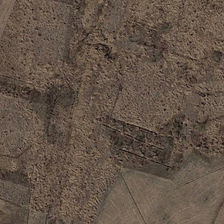}}}\medskip
    \end{minipage}
    \begin{minipage}[c]{\figwidth}
        \centering
        \fcolorbox{wrong}{wrong}{\centerline{\includegraphics[height=\figheight]{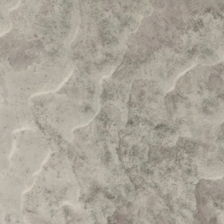}}}\medskip
    \end{minipage}

%JDSH T2I
    \begin{minipage}[c]{\figwidth}
        \centering
        \centerline{
        \parbox[t][0.6in][c]{0.6in}{JDSH \cite{liu2020joint}}
        }
    \end{minipage}
    \begin{minipage}[c]{\figwidth}
        \tiny
        \color{neutral}
        \centerline{
        \parbox[t][0.6in][t]{0.6in}{The wasteland has both striped and radial texture}
        }
    \end{minipage}
    \begin{minipage}[c]{\figwidth}
        \centering
         \fcolorbox{neutral}{neutral}{\centerline{\includegraphics[height=\figheight]{images/retrievals/bareland_45.jpg}}}\medskip
    \end{minipage}
    \begin{minipage}[c]{\figwidth}
        \centering
        \fcolorbox{correct}{correct}{\centerline{\includegraphics[height=\figheight]{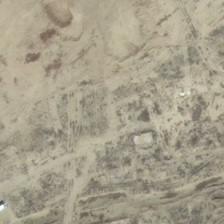}}}\medskip
    \end{minipage}
    \begin{minipage}[c]{\figwidth}
        \centering
        \fcolorbox{correct}{correct}{\centerline{\includegraphics[height=\figheight]{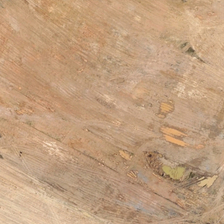}}}\medskip
    \end{minipage} 
    \begin{minipage}[c]{\figwidth}
        \centering
        \fcolorbox{correct}{correct}{\centerline{\includegraphics[height=\figheight]{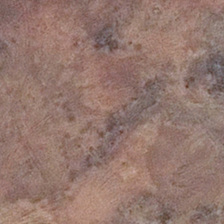}}}\medskip
    \end{minipage}
    \begin{minipage}[c]{\figwidth}
        \centering
        \fcolorbox{correct}{correct}{\centerline{\includegraphics[height=\figheight]{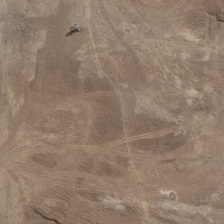}}}\medskip
    \end{minipage}
    \begin{minipage}[c]{\figwidth}
        \centering
        \fcolorbox{correct}{correct}{\centerline{\includegraphics[height=\figheight]{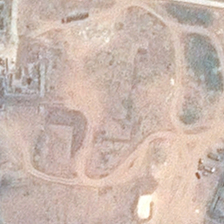}}}\medskip
    \end{minipage}
    \begin{minipage}[c]{\figwidth}
        \centering
        \fcolorbox{correct}{correct}{\centerline{\includegraphics[height=\figheight]{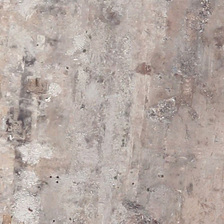}}}\medskip
    \end{minipage}
    \begin{minipage}[c]{\figwidth}
        \centering
        \fcolorbox{wrong}{wrong}{\centerline{\includegraphics[height=\figheight]{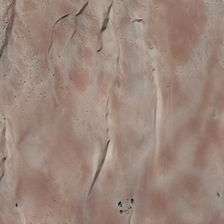}}}\medskip
    \end{minipage}

%DUCH T2I
    \begin{minipage}[c]{\figwidth}
        \centering
        \centerline{
        \parbox[t][0.6in][c]{0.6in}{DUCH \small\textit{(proposed)}}
        }
    \end{minipage}
    \begin{minipage}[c]{\figwidth}
        \tiny
        \color{neutral}
        \centerline{
        \parbox[t][0.6in][t]{0.6in}{The wasteland has both striped and radial texture}
        }
    \end{minipage}
    \begin{minipage}[c]{\figwidth}
        \centering
        \fcolorbox{neutral}{neutral}{\centerline{\includegraphics[height=\figheight]{images/retrievals/bareland_45.jpg}}}\medskip
    \end{minipage}
    \begin{minipage}[c]{\figwidth}
        \centering
        \fcolorbox{correct}{correct}{\centerline{\includegraphics[height=\figheight]{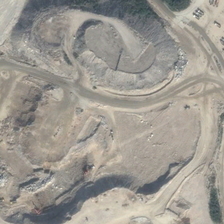}}}\medskip
    \end{minipage}
    \begin{minipage}[c]{\figwidth}
        \centering
        \fcolorbox{correct}{correct}{\centerline{\includegraphics[height=\figheight]{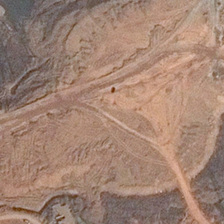}}}\medskip
    \end{minipage} 
    \begin{minipage}[c]{\figwidth}
        \centering
        \fcolorbox{correct}{correct}{\centerline{\includegraphics[height=\figheight]{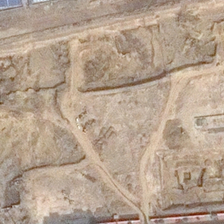}}}\medskip
    \end{minipage}
    \begin{minipage}[c]{\figwidth}
        \centering
        \fcolorbox{correct}{correct}{\centerline{\includegraphics[height=\figheight]{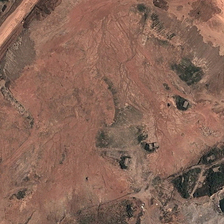}}}\medskip
    \end{minipage}
    \begin{minipage}[c]{\figwidth}
        \centering
        \fcolorbox{correct}{correct}{\centerline{\includegraphics[height=\figheight]{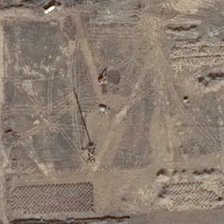}}}\medskip
    \end{minipage}
    \begin{minipage}[c]{\figwidth}
        \centering
        \fcolorbox{correct}{correct}{\centerline{\includegraphics[height=\figheight]{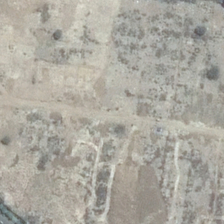}}}\medskip
    \end{minipage}
    \begin{minipage}[c]{\figwidth}
        \centering
        \fcolorbox{correct}{correct}{\centerline{\includegraphics[height=\figheight]{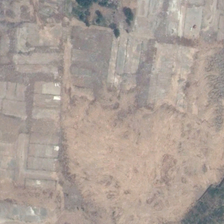}}}\medskip
    \end{minipage}

    \caption{$T \to I$ retrieval results obtained when $B=64$ for RSICD dataset.}
    \label{fig:retrieval_t2i}
\end{figure*}

\noindent\emph{Results for the UCMerced dataset:} Table~\ref{tab:retrieval-performance-ucm} shows the corresponding cross-modal retrieval performances on the UCMerced dataset. By analyzing the table, one can see that the mAP@20 increases for all methods by increasing the number of the hash bits for image-to-text and text-to-image retrieval tasks. As an example, by increasing the value of $B$ from 16 to 128, the proposed DUCH improved in the mAP@20 more than 10\% and 12\% in the image-to-text and text-to-image retrieval tasks, respectively. Furthermore, the proposed DUCH method sharply outperformed all the unsupervised baselines, especially when hash codes with short lengths are considered. As an example, for $I \to T$ retrieval task when $B=16$ the proposed DUCH obtained about 8\% and 30\% higher scores than DJSRH and JDSH, respectively. Similarly, for $T \to I$ retrieval task when $B=32$, the proposed DUCH outperformed DJSRH and JDSH with 10\% and 6\% higher mAP@20 scores, respectively. In the UCMerced dataset, the supervised method CPAH obtained the best performance in terms of mAP@20. As an example, when $B$=64 for $I \to T$ and $T\to I$ retrieval tasks, the supervised CPAH method obtained about 5\% and 7\% higher scores than DUCH, respectively. The high performance of the supervised CPAH is obtained at the cost of using labeled samples for hash code learning. Since the UCMerced dataset is relatively small, the use of labeled samples can significantly boost the performance of the supervised methods.

\noindent\emph{Results for the RSICD dataset:}
Table~\ref{tab:retrieval-performance-rsicd} shows the cross-modal retrieval performances for image-to-text and text-to-image retrieval tasks on the RSICD dataset. One can see that similar to the UCMerced results, the mAP@20 for all methods increases as the value for $B$ grows. As an example, by increasing the value of $B$ from 16 to 128, the proposed DUCH and CPAH improved  the mAP@20 more than 12\% and 25\% in the $T \to I$ retrieval task, respectively. By analyzing the results, one can observe that the proposed DUCH method sharply outperformed almost all the supervised and unsupervised baselines when $B>16$. As an example, for $I \to T$ retrieval task when $B=128$ the proposed DUCH obtained about 12\% and 10\% higher scores than CPAH (supervised) and DJSRH (unsupervised), respectively. Similarly, for $T \to I$ retrieval task when $B=64$, the proposed DUCH outperformed JDSH and DJSRH with 2\% and 12\% higher mAP@20 scores, respectively. In the RSICD dataset, the to the proposed DUCH significantly better performance compared to supervised method CPAH in terms of mAP@20. As an example, when $B$=64 for $I \to T$ and $T\to I$ retrieval tasks, the supervised CPAH method obtained about 20\% and 15\% lower scores than DUCH, respectively. This is mainly due to the larger scale of the RSICD dataset compared to the UCMerced. This demonstrates that we can fill the performance gap without any requirement on labeled samples by considering more aligned text and image samples to train our unsupervised DUCH. We further analyze the precision of DUCH for the number $K$ of retrieved images in comparison with the baselines. Precision versus the number of retrieved images when $B$=64 are shown in Fig. \ref{fig:precision-at-k}. From the figure, one can observe that the precision obtained by unsupervised DUCH (in all $K$ values) provides higher precision associated with each $K$ in comparison to all the supervised and unsupervised baseline methods. This indicates that the proposed DUCH in compared to the other unsupervised methods can learn to map semantic information into discriminative hash codes without using the label information. 

Fig. \ref{fig:retrieval_i2t}  shows an example of cross-modal retrieval results for image-to-text from RSICD by CPAH, DJSRH, JDSH and the proposed DUCH when the query image is an airport area including many buildings and airplanes. The related captions to the query image and the ordered retrieved captions obtained by different methods are presented in the figure. By analyzing the figure, one can observe that the classes of \textit{Airplane} and \textit{Airport} are very prominent in all retrieved captions by JDSH and the proposed DUCH, while CPAH and DJSRH confuse the airport building with a church retrieving unrelated captions to the query image. When DUCH is compared with JDSH, both methods retrieve relevant captions, while the proposed method retrieves semantically more relevant captions. As an example, 3rd retrieved caption by JDSH failed to estimate the number of airplanes, while the proposed DUCH consistently demonstrated the quantity correctly. The qualitative cross-modal retrieval results for text-to-image retrieval from RSICD are shown in Fig. \ref{fig:retrieval_t2i} when a query caption includes ``wasteland'' and ``radial texture''. The related image to the query caption and the ordered retrieved images obtained by CPAH, DJSRH, JDSH and the proposed DUCH are presented in the figure. The figure shows that all of the compared methods are confused between ``deserts'' and ``barren lands'', while the proposed DUCH  always retrieved semantically similar images. As an example, CPAH provides similar images to the query caption only at the retrieved orders of 2, 3 and 4. By analyzing the results presented in Fig. \ref{fig:retrieval_i2t} and Fig. \ref{fig:retrieval_t2i} one can observe that the proposed DUCH brings more consistency over intra-modality than the compared baseline methods. The reason is that the proposed DUCH: i) uses a text augmentation method that always preserves the semantic similarity between the original caption and the augmented caption; and ii) considers the intra-modality contrastive loss term enforcing the intra-modality consistency. These results are also confirmed for the UCMerced archive (not reported for space constraints). 

%To demonstrate the effectiveness of the learned representations, we used t-distributed stochastic neighbor embedding (t-SNE) to visualize learned feature spaces of the feature extraction and hashing modules. Figures \ref{fig:tsne-img} and \ref{fig:tsne-txt} show the t-SNE visualization of extracted features and binary codes obtained in the RSICD dataset for image and text modalities, respectively. Smaller Hamming distance between samples results in a smaller distance in t-SNE visualization. NEXT TWO SENTENCES ARE NOT CONVINGING- NO IDEA WHAT SEMANTIC SPACE AND REPRESENTATIVE FEATURE SPACES ARE
%By analysing the figures, one can observe that the hashing process simultaneously increases the gap between different sample groups and reduces the distance between samples of the same class in the embedding space. This demonstrates that the proposed DUCH could learn representative feature spaces and hashing functions. The same relative behavior is also observed in the results obtained by using the UCMerced dataset (not reported for space constraints). 

\section{Conclusion}
\label{sec:conclusion}
 
In this paper, a novel deep unsupervised contrastive hashing (DUCH) method has been proposed for cross-modal image-text retrieval in RS. The proposed DUCH includes two main modules: 1) feature extraction module; and 2) hashing module. The feature extraction module consists of two modality-specific encoders for image and text modalities. Within the hashing module, we have introduced a multi-term loss function for unsupervised learning of cross-modal feature representations. In detail, a contrastive objective function that considers both inter- and intra-modal similarities has been introduced. Furthermore, an adversarial loss term has been presented to enforce the generation of modality-invariant representations. We have also introduced a novel rule-based text augmentation method based on replacement of related synonyms of the words in the captions. Experimental results obtained on two benchmark archives (those that include RS image-text pairs) show the superiority of the proposed DUCH compared to the state-of-the-art unsupervised cross-modal retrieval methods, while providing similar results with respect to a supervised method that requires a high number of labeled training samples. In addition, we have analyzed the effect of different data augmentation methods on the performance of DUCH to demonstrate the importance of applying appropriate data augmentation methods for unsupervised contrastive learning. We show that there is a best-performing value for the hyperparameters of each augmentation method. Specifically, the color jittering and Gaussian blur augmentation methods are very sensitive to the hyperparameters selection since they can eliminate information when applied to RS images affecting the retrieval performance. As a final remark we would like to note that through the experimental results we also observed that the intra-modal contrastive loss is less effective when the deep features are obtained from a modality-specific encoder pre-trained on a different domain (e.g., CV images). To address this problem, as a future work, we plan to apply end-to-end training of the feature extraction module by fine-tuning the modality-specific encoders.

\appendices

% \section*{Appendix}

\section*{Acknowledgments}
This work is funded by the European Research Council (ERC) through the ERC-2017-STG BigEarth Project under Grant 759764 and by the German Ministry for Education and Research as BIFOLD - Berlin Institute for the Foundations of Learning and Data (01IS18025A). The authors would like to thank Genc Hoxha from University of Trento for the initial discussions on cross-modal image retrieval for remote sensing.

\bibliography{ref}
\bibliographystyle{ieeetr}

\end{document}